\pdfoutput=1
\documentclass[11pt]{article}

\usepackage{acl}

\usepackage{times}
\usepackage{latexsym}
\usepackage[T1]{fontenc}
\usepackage[utf8]{inputenc}
\usepackage{microtype}
\usepackage{inconsolata}
\usepackage{amsmath,amssymb,bm}
\usepackage{graphicx}
\usepackage{booktabs,multirow,array,longtable,tabularx}
\usepackage{enumitem}
\usepackage{caption,subcaption,float}
\usepackage[ruled,linesnumbered]{algorithm2e}
\usepackage{array}
\usepackage{multirow}
\usepackage{tabularx}
\newcommand{\Norm}{\operatorname{Norm}}
\newcommand{\LN}{\operatorname{LN}}
\newcommand{\BN}{\operatorname{BN}}

\newcommand{\ind}{\mathbb{I}}
\newcommand{\R}{\mathbb{R}}
\newcommand{\meanstd}[2]{#1{\scriptsize\,\(\pm #2\)}}

\title{RAEGNet: Relation-Aware Evidence Graph Network for Harm-Aware Multimodal Fake News Detection}

\author{
\textbf{Wenbin Shen}$^{*1}$ \quad
\textbf{Guoxuan Qin}$^{1}$ \quad
\textbf{Guangxu Yao}$^{1}$ \quad
\textbf{Baodong Wang}$^{1}$ \quad
\textbf{Yuanbo Rui}$^{1}$ \\
\textbf{Zhongjie Ba}$^{2}$ \quad
\textbf{Zhichao Lian}$^{*1,3}$ \\[4pt]
$^{1}$Nanjing University of Science and Technology \\[2pt]
$^{2}$Zhejiang University \\[2pt]
$^{3}$University of Chinese Academy of Sciences \\[3pt]
\texttt{shenwenbin@njust.edu.cn;\quad
lzcts@163.com}
}

\begin{document}
\maketitle

\begin{abstract}
Existing multimodal fake news detection methods often introduce external information to assist detection. However, most of them rely on entity-level retrieval and are therefore prone to introducing event-irrelevant noise. Meanwhile, existing methods mainly focus on improving overall performance and do not account for differences in the degree of harm posed by different instances of fake news. To address these limitations, we design an \textbf{E}vent-\textbf{L}evel \textbf{E}vidence \textbf{R}etrieval \textbf{F}ramework (\textbf{ELERF}) and propose a \textbf{R}elation-\textbf{A}ware \textbf{E}vidence \textbf{G}raph \textbf{Net}work (\textbf{RAEGNet}). ELERF retrieves external evidence based on the complete event semantics of a news item. RAEGNet constructs a directed graph that incorporates news-evidence stance relations and evidence-evidence interaction relations, and introduces a conditional-harm branch to jointly model authenticity and potential harm. Experimental results demonstrate that RAEGNet outperforms multiple baseline methods across all evaluated metrics on Weibo-21, Fakeddit, and our self-constructed SSS dataset.
\end{abstract}

\section{Introduction}
The rapid development of social media has greatly accelerated the dissemination of multimodal information, while also providing fertile ground for fake news, posing severe threats to public security \citep{shu2017fake, zhou2020survey}. In the real-world cyber ecosystem, different instances of fake news vary substantially in their potential destructive impact. For example, political deepfakes and public-health rumors can cause far greater social harm than poorly fabricated entertainment gossip \citep{shu2020fakenewsnet, nan2021mdfend}. Therefore, multimodal fake news detection not only requires the foundational capability to accurately discern the authenticity of information but more urgently needs to prioritize the perception of highly harmful news so as to minimize the substantive adverse effects of misinformation on society \citep{liu2018early}.

\begin{figure}[t]
    \centering
    \includegraphics[width=\columnwidth]{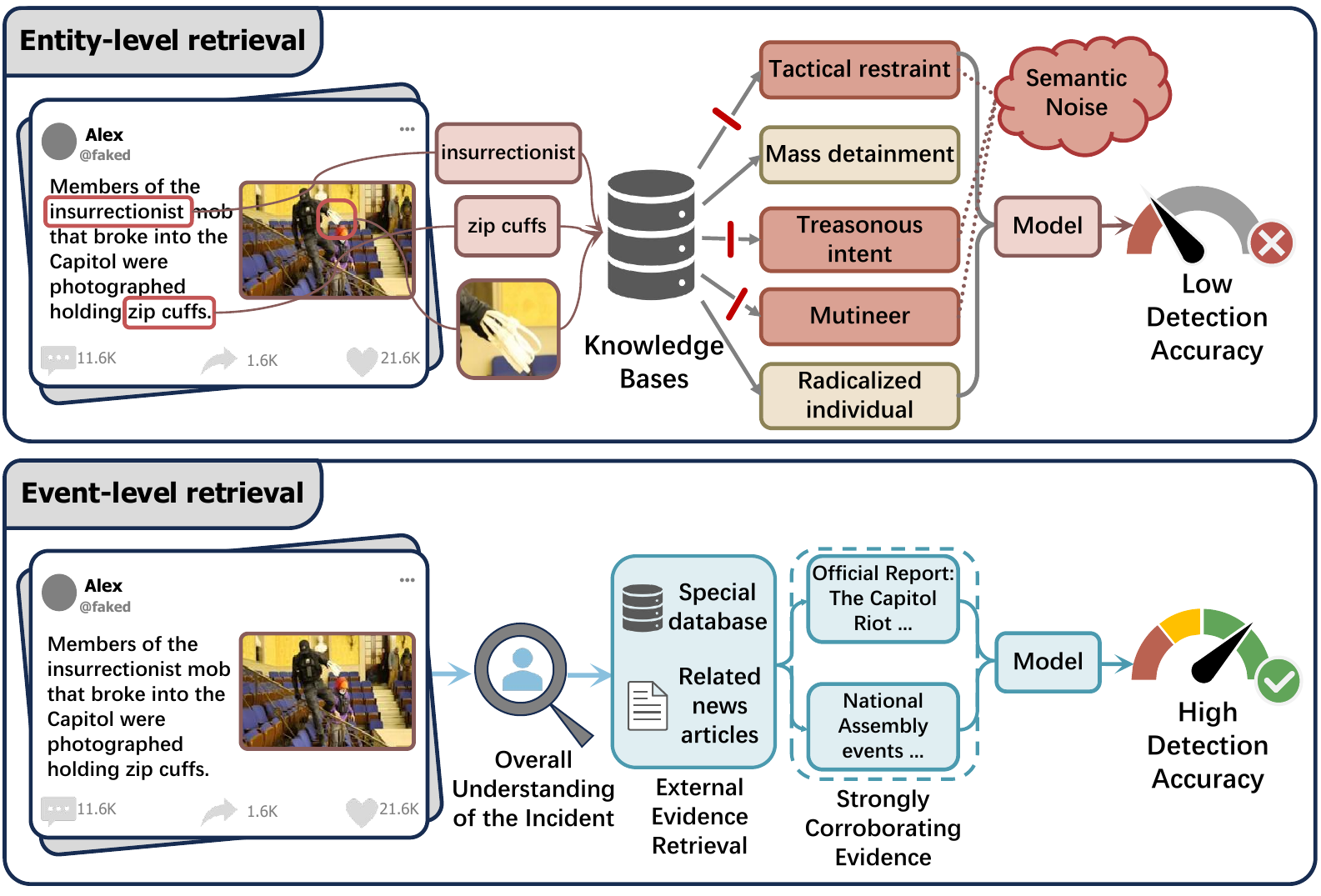}
    \caption{Comparison between entity-level retrieval and our event-level retrieval. }
    \label{fig:retrieval_comparison}
\end{figure}

Although existing multimodal fake news detection techniques have made substantial progress \citep{qi2019exploiting, zhou2020similarity, silva2021embracing, singhal2019spotfake, wang2024fake}, two core issues remain. First, regarding the incorporation of external knowledge, existing methods mostly rely on entity-level retrieval \citep{popat2018declare, augenstein2019multifc} and knowledge-graph linking \citep{pan2018content, dun2021kan, kim2023factkg}. Such methods easily deviate from the holistic semantics of the news event and introduce substantial semantic noise unrelated to the central event \citep{fu2023kg}. Second, most existing studies formulate the objective as a simple binary classification task and focus only on improving aggregate performance, ignoring the significant variance in the degree of harmfulness among samples. In real-world settings, the social cost of failing to detect high-harm fake news is much higher than that of misclassifying an ordinary rumor. Although some studies have attempted to grade the severity of fake news \citep{reddy2023fndsd, pillai2024hierarchical}, they are mostly limited to text-only settings or separate, simple assessments. How harm judgments can be intrinsically integrated into a multimodal detection framework remains insufficiently explored.

To address these challenges, we design the Event-Level Evidence Retrieval Framework (ELERF) and propose the Relation-Aware Evidence Graph Network (RAEGNet). ELERF uses the complete news text as the query and obtains strongly relevant external evidence from a predefined source pool through retrieval, filtering, and stance quantification. Based on the stance of each piece of evidence toward the news and the semantic relations among evidence items, RAEGNet constructs an evidence graph containing nine types of directed edges. In addition, by screening and integrating existing public data, we construct the SSS multimodal fake news dataset. SSS combines six public data sources and, after validity checks, sample deduplication, semantic filtering, and manual review, contains 7,997 Chinese and English multimodal news samples. Each sample is provided with a harm label, external evidence, and the corresponding evidence stance scores, thereby supporting authenticity detection, harm analysis, and relational evidence reasoning.

The main contributions of this paper are as follows:
\begin{itemize}
    \item We propose an event-level evidence retrieval framework that reduces semantic noise and provides external evidence strongly related to the complete news event.
    \item We propose a relation-aware evidence graph that jointly models the relation types and relation strengths of news-evidence and evidence-evidence interactions.
    \item We propose a harm-aware joint optimization objective that jointly learns authenticity classification, conditional-harm estimation, and ordinal harm consistency.
    \item We construct the SSS multimodal fake news dataset, which provides high-quality Chinese and English news samples together with external evidence.
\end{itemize}

\section{Related Work}
\subsection{Multimodal Fake News Detection}
The development of multimodal fake news detection methods closely follows the iteration of feature extraction technologies. Early studies primarily utilized CNNs and RNNs to extract visual and textual features, fusing them through simple concatenation or attention mechanisms \citep{jin2017multimodal, yang2018ti}. With breakthroughs in pre-training technologies, the field has entered an era of fine-grained cross-modal interaction. For instance, CAFE \citep{chen2022cross} utilizes fuzzy reasoning to quantify cross-modal ambiguity, InfoSurgeon \citep{fung2021infosurgeon} constructs a consistency-checking graph by aligning text entities with visual regions, and CCGN \citep{cui2025ccgn} directly captures the inconsistency between multimodal contents within a contrastive learning space. These methods significantly enhance the capability to capture image-text contradictions. However, restricted by the information inherent in the news itself, they struggle to handle deeply disguised fake news.

\subsection{Knowledge-Enhanced Detection Methods}
Introducing external knowledge becomes the key path to breaking through the limitations of internal information, and it can be mainly categorized into three types. The first category relies on static knowledge graphs. For instance, CompareNet \citep{hu2021compare} aligns news texts with KGs for entity alignment to determine authenticity, while AKA-Fake \citep{zhang2024reinforced} utilizes reinforcement learning to adaptively retrieve relevant knowledge subgraphs. The second category is based on open-domain retrieval. For instance, MUSER \citep{liao2023muser} mimics the logical process of human fact-checking and automatically collects key evidence through a multi-step retrieval mechanism. The third category is the paradigm based on LLMs and multi-agent systems, which treats LLMs as powerful implicit knowledge bases, reasoning engines, and orchestration centers. Typically, FKA-Owl \citep{liu2024fka} leverages LLMs to mine implicit commonsense, while RAMA \citep{yang2025rama} proposes a retrieval-augmented multi-agent framework. However, the majority of existing knowledge enhancement strategies still heavily rely on entity-level or shallow semantic matching, which makes them prone to introducing irrelevant noise and consequently degrading the overall performance.

\begin{figure*}[t]
    \centering
    \includegraphics[width=\textwidth]{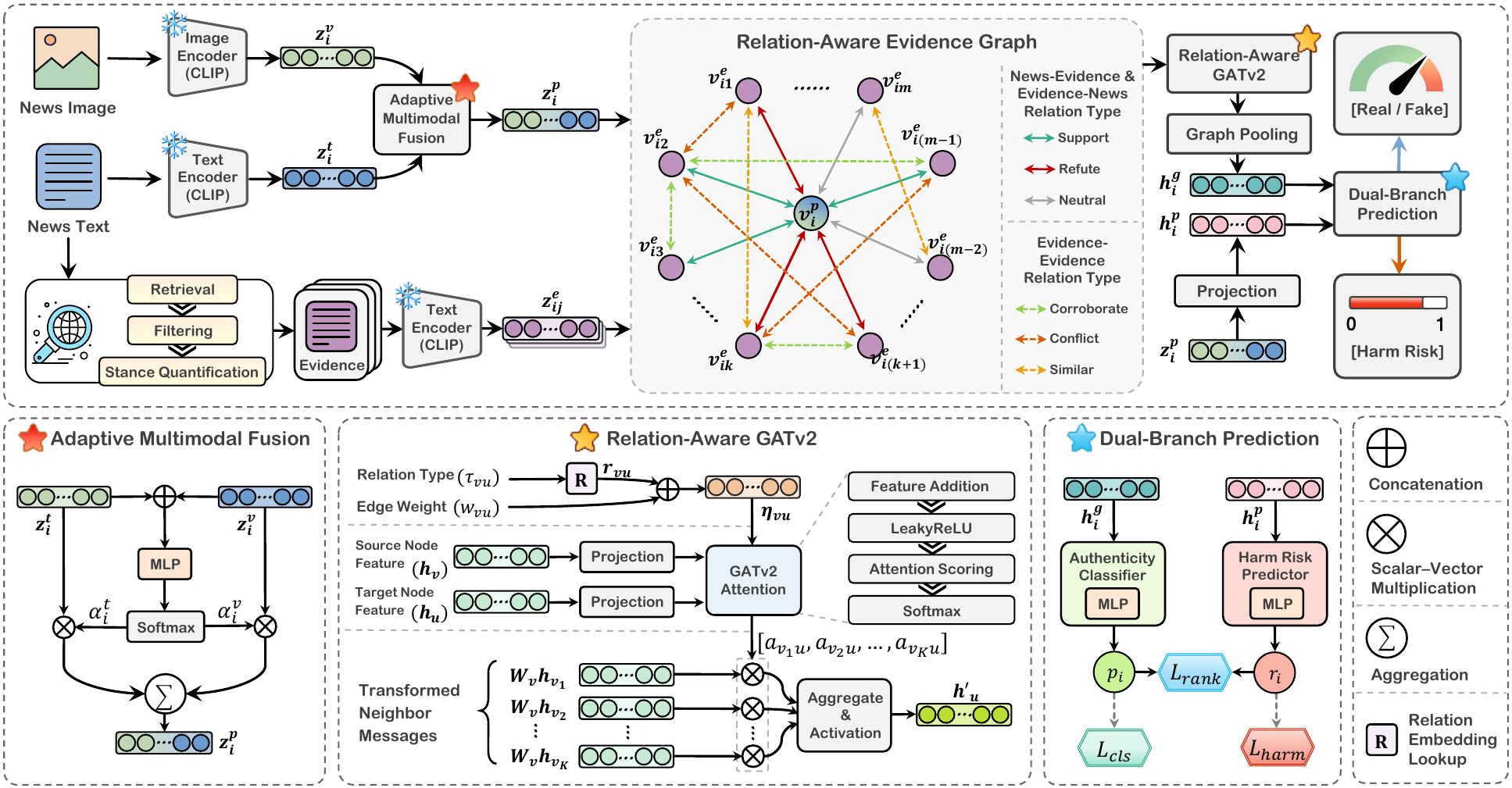}
    \caption{The overall architecture of RAEGNet. }
    \label{fig:overall_architecture}
\end{figure*}

\section{Methodology}
\subsection{Overview}

As shown in Figure~\ref{fig:overall_architecture}, RAEGNet first jointly encodes the news text, image, and evidence. It then constructs a relation-aware evidence graph for authenticity reasoning. At the same time, it uses the harm-risk branch to estimate conditional harm.

\subsection{Event-Level Evidence Retrieval}
To reduce irrelevant noise introduced by entity-level retrieval, we design an event-level evidence retrieval framework comprising the following three steps.

\begin{itemize}
    \item \textbf{Retrieval.} For the $i$-th news, we use the complete news text as an event-level query and retrieve the top $N$ webpage snippets together with their metadata as candidate evidence.
    \item \textbf{Filtering.} Candidate webpages with explicit verdict-style information are automatically removed. The remaining candidates are further screened by an LLM to discard evidence containing implicit label leakage.
    \item \textbf{Stance quantification}. For each leakage-free candidate evidence item, the LLM evaluates a continuous stance score $\widetilde q_{ij}\in[0,1]$, which is then mapped to a discrete stance score $q_{ij}\in\{0.1,0.3,0.5,0.7,0.9\}$. The final evidence set is $\mathcal E_i=\{(E_{ij},q_{ij})\}_{j=1}^{m_i}$, where $m_i\leq N$.
\end{itemize}

\subsection{Feature Encoding and Fusion}
Let $T_i$ and $I_i$ denote the text and image of the $i$-th news item, respectively, and let $f_t(\cdot)$ and $f_v(\cdot)$ denote the text and image encoders. The normalized representations of the news text, news image, and $j$-th evidence are $\mathbf z_i^t=\Norm(f_t(T_i)),\mathbf z_i^v=\Norm(f_v(I_i))$ and 
$\mathbf z_{ij}^e=\Norm(f_t(E_{ij}))$, respectively. Here,  $\Norm(\mathbf x)=\mathbf x/\|\mathbf x\|_2$.

After obtaining the modality representations in the unified space, the fused news feature is computed as
\begin{equation}
\mathbf z_i^p=\alpha_i^t\mathbf z_i^t+\alpha_i^v\mathbf z_i^v,
\end{equation}
where $\alpha_i^t$ and $\alpha_i^v$ are the weights of the news text feature $\mathbf z_i^t$ and image feature $\mathbf z_i^v$, respectively. They are computed as
\begin{equation}
\boldsymbol\alpha_i=\operatorname{softmax}\!\left(
\mathbf W_2\phi\!\left(\mathbf W_1[\mathbf z_i^t;\mathbf z_i^v]+\mathbf b_1\right)+\mathbf b_2
\right),
\end{equation}
where $\mathbf W_1,\mathbf W_2,\mathbf b_1,$ and $\mathbf b_2$ are learnable parameters of the gating network; $\phi(\cdot)$ is the ReLU activation function; and $\boldsymbol\alpha_i=[\alpha_i^t,\alpha_i^v]$, with $\alpha_i^t,\alpha_i^v\geq0$ and $\alpha_i^t+\alpha_i^v=1$.

\subsection{Relation-Aware Evidence Graph Network}
For the $i$-th news item, we construct a graph $\mathcal G_i=(\mathcal V_i,\mathcal A_i)$. Its node set is $\mathcal V_i=\{v_i^p\}\cup\{v_{i1}^e,v_{i2}^e,\ldots,v_{im}^e\}.$ Here, $v_i^p$ denotes the central news node, whose original feature is $\mathbf z_i^p$, and $v_{ij}^e$ denotes the evidence node corresponding to evidence $E_{ij}$, whose original feature is $\mathbf z_{ij}^e$. For any node $u\in\mathcal V_i$, let $\mathbf z_u$ denote its original feature. Its initial node state is obtained through a shared projection:
\begin{equation}
\mathbf x_u^{(0)}=\phi\!\left(
\LN(\mathbf W_e\mathbf z_u+\mathbf b_e)
\right),
\end{equation}
where $\mathbf W_e$ and $\mathbf b_e$ are the shared projection parameters.

After node initialization, news--evidence relations are constructed according to the semantic relevance and stance direction between the news text and the external evidence. First, the nonnegative cosine relevance between the news text and the evidence text is computed as
\begin{equation}
c_{ij}=\max\!\left(0,\cos(\mathbf z_i^t,\mathbf z_{ij}^e)\right).
\end{equation}
Then, based on the stance score $q_{ij}$, a support edge $P\!\rightarrow\!E^+$, a refutation edge $P\!\rightarrow\!E^-$, or a neutral edge $P\!\rightarrow\!E^0$ is constructed. Let the support threshold, refutation threshold, and neutral-neighborhood width be $\delta_+$, $\delta_-$, and $\delta_0$, respectively, satisfying $0<\delta_-<\frac{1}{2}<\delta_+<1, 0<\delta_0\leq\min\left\{\frac{1}{2}-\delta_-,\delta_+-\frac{1}{2}\right\}.$

The relation type $\tau_{ij}$ and edge weight $w_{ij}$ are defined as
\begin{equation}
(\tau_{ij},w_{ij})=
\begin{cases}
(P\!\rightarrow\!E^+,\;q_{ij}c_{ij}), & q_{ij}\geq\delta_+,\\
(P\!\rightarrow\!E^-,\;(1-q_{ij})c_{ij}), & q_{ij}<\delta_-,\\
(P\!\rightarrow\!E^0,\;q_{ij}c_{ij}), & \mu_0<\delta_0,
\end{cases}
\end{equation}
where $\mu_0=|q_{ij}-1/2|$.

A reverse edge with the same weight is added for every valid edge, denoted by $E\!\rightarrow\!P^+$, $E\!\rightarrow\!P^-$, and $E\!\rightarrow\!P^0$, respectively.

We next model relations among evidence items. For any two evidence nodes $v_{ij}^e$ and $v_{ik}^e$, the relevance between their textual representations is computed as
\begin{equation}
c_{ijk}=\max\!\left(0,\cos(\mathbf z_{ij}^e,\mathbf z_{ik}^e)\right),
\end{equation}
and is used as the weight of the evidence--evidence edge.

The news--evidence relations are then mapped to stance groups $g_{ij}\in\{\mathrm{support},\mathrm{refute},\mathrm{neutral}\}$. Let $\kappa_{\mathrm{cor}}$, $\kappa_{\mathrm{con}}$, and $\kappa_{\mathrm{sim}}$ denote the relevance thresholds for corroboration, conflict, and similarity, respectively. Evidence relations are determined in the following order of priority:

\begin{enumerate}[label=\arabic*.,leftmargin=2em]
  \item If $g_{ij}=g_{ik}\in\{\mathrm{support},\mathrm{refute}\}$ and $c_{ijk}\geq\kappa_{\mathrm{cor}}$, bidirectional corroboration edges $E\!\leftrightarrow\!E^{\mathrm{cor}}$ are added.
  \item If $\{g_{ij},g_{ik}\}=\{\mathrm{support},\mathrm{refute}\}$ and $c_{ijk}\geq\kappa_{\mathrm{con}}$, bidirectional conflict edges $E\!\leftrightarrow\!E^{\mathrm{con}}$ are added.
  \item If neither of the first two conditions is satisfied and $c_{ijk}\geq\kappa_{\mathrm{sim}}$, bidirectional similarity edges $E\!\leftrightarrow\!E^{\mathrm{sim}}$ are added.
\end{enumerate}

The complete relation set $\mathcal T$ therefore consists of three types of news--evidence edges, three types of evidence--news edges, and three types of evidence--evidence edges.

For any directed edge $(v,u)\in\mathcal A_i$, let $\tau_{vu}$ denote its relation type and $w_{vu}$ its edge weight. We embed the nine relation types with a learnable matrix $\mathbf R\in\R^{|\mathcal T|\times d_r}$ and obtain $\mathbf r_{vu}=\mathbf R[\tau_{vu}]\in\R^{d_r}$. The edge feature is then $\boldsymbol\eta_{vu}=[w_{vu};\mathbf r_{vu}]$, which keeps both relation strength and relation category.

We use GATv2 \cite{brody2022how} as the message-passing operator. Let $\mathcal N(u)=\{v:(v,u)\in\mathcal A_i\}$ denote the in-neighbors of node $u$. At layer $\ell$, the attention score for edge $(v,u)$ is computed from the target-node state, source-node state, and edge feature:
\begin{equation}
e_{vu}^{(\ell)}=\mathbf a_{\ell}^{\top}
\psi\!\left(
\mathbf W_t^{(\ell)}\mathbf x_u^{(\ell)}+
\mathbf W_s^{(\ell)}\mathbf x_v^{(\ell)}+
\mathbf W_{\eta}^{(\ell)}\boldsymbol\eta_{vu}
\right).
\end{equation}
where $\mathbf W_t^{(\ell)}$, $\mathbf W_s^{(\ell)}$, $\mathbf W_{\eta}^{(\ell)}$, and $\mathbf a_\ell$ are layer-specific parameters, and $\psi(\cdot)$ is the LeakyReLU activation function. The scores of all incoming edges to node $u$ are normalized by softmax:
\begin{equation}
a_{vu}^{(\ell)}=
\frac{\exp(e_{vu}^{(\ell)})}
{\sum_{k\in\mathcal N(u)}\exp(e_{ku}^{(\ell)})}.
\end{equation}
The next-layer state of node $u$ is obtained by weighted aggregation:
\begin{equation}
\mathbf x_u^{(\ell+1)}=\phi\!\left(
\sum_{v\in\mathcal N(u)}a_{vu}^{(\ell)}\mathbf W_m^{(\ell)}\mathbf x_v^{(\ell)}
\right),
\end{equation}
where $\mathbf W_m^{(\ell)}$ is the message transformation matrix.

After $L$ propagation layers, mean pooling and layer normalization produce the evidence-enhanced news representation
\begin{equation}
\mathbf h_i^g=\LN\!\left(
\frac{1}{|\mathcal V_i|}\sum_{u\in\mathcal V_i}\mathbf x_u^{(L)}
\right).
\end{equation}
Finally, based on $\mathbf h_i^g$, the classifier outputs the probability that the news item is fake:
\begin{equation}
p_i=\sigma\!\left(
\mathbf w_o^{\top}\mathcal D\!\left(
\phi\!\left(\BN(\mathbf W_c\mathbf h_i^g+\mathbf b_c)\right)
\right)+b_o
\right),
\end{equation}
where $\mathbf W_c,\mathbf b_c,\mathbf w_o,$ and $b_o$ are classifier parameters; $\BN$ denotes batch normalization; $\mathcal D(\cdot)$ denotes dropout; and $\sigma$ is the sigmoid function. Given a classification threshold $\theta_y$, the predicted label is $\hat y_i=\ind(p_i\geq\theta_y)$.

\subsection{Joint Optimization with Harm Risk}
For the $i$-th news item, let $y_i\in\{0,1\}$ denote its authenticity label, where $y_i=1$ indicates fake news. The conditional harm label $h_i\in(0,1)$ represents the potential impact of the content if it were false and disseminated. In annotation, harm is collected as five ordered levels and represented by $\mathcal H=\{a_1,a_2,a_3,a_4,a_5\}=\{0.1,0.3,0.5,0.7,0.9\}$. The authenticity branch outputs the fake-news probability $p_i$ through the relation-aware evidence graph. The harm-risk branch takes the news content representation as input and first obtains a hidden representation
\begin{equation}
\mathbf h_i^p=\LN(\mathbf W_p\mathbf z_i^p+\mathbf b_p),
\end{equation}
after which it predicts the conditional harm score
\begin{equation}
r_i=\sigma\!\left(
\mathbf w_r^{\top}\phi(\mathbf W_r\mathbf h_i^p+\mathbf b_r)+\beta_r
\right),
\end{equation}
where $\mathbf W_p,\mathbf b_p,\mathbf W_r,\mathbf b_r,\mathbf w_r,$ and $\beta_r$ are parameters of the harm branch.

We jointly optimize authenticity classification and conditional-harm modeling through three complementary objectives. First, binary cross-entropy is used for authenticity classification:
\begin{equation}
\mathcal L_{\mathrm{cls}}=-\frac{1}{B}\sum_{i=1}^{B}
\left[y_i\log p_i+(1-y_i)\log(1-p_i)\right],
\end{equation}
where $B$ is the mini-batch size.

Second, mean squared error is used to constrain the harm-risk branch to fit the conditional harm labels:
\begin{equation}
\mathcal L_{\mathrm{harm}}=\frac{1}{B}\sum_{i=1}^{B}(r_i-h_i)^2.
\end{equation}
This objective provides pointwise supervision for conditional-harm estimation.

Finally, a harm-aware ranking constraint is introduced to preserve the ordinal structure of conditional harm. We define the ordered sample pairs in a mini-batch as
\begin{equation}
\mathcal P_B=\{(i,j):h_i>h_j\}.
\end{equation}
For each ordered pair, the relative harm difference is normalized as
$\Delta_{ij}=(h_i-h_j)/(\max(\mathcal H)-\min(\mathcal H))$,
and the ranking objective is defined as
\begin{equation}
\mathcal L_{\mathrm{rank}}=
\frac{1}{|\mathcal P_B|}
\sum_{(i,j)\in\mathcal P_B}
\left[
\gamma\Delta_{ij}-(r_i-r_j)
\right]_+^2,
\end{equation}
where $[\cdot]_+=\max(0,\cdot)$ and $\gamma\geq0$ controls the ranking margin. When $\mathcal P_B$ is empty, $\mathcal L_{\mathrm{rank}}$ is set to 0. This objective encourages consistency between the relative ordering of predicted harm scores and the annotated harm levels.

Combining the three objectives, the final optimization objective is
\begin{equation}
\mathcal L= \mathcal L_{\mathrm{cls}}+
\lambda_{\mathrm{harm}}\mathcal L_{\mathrm{harm}}+
\lambda_{\mathrm{rank}}\mathcal L_{\mathrm{rank}},
\end{equation}
where $\lambda_{\mathrm{harm}}$ and $\lambda_{\mathrm{rank}}$ are nonnegative weights. The joint objective enables the model to learn authenticity discrimination while simultaneously capturing both the absolute magnitude and ordinal structure of conditional harm. In this way, authenticity prediction and harm modeling are optimized within a unified framework.

\section{Experiments}

\subsection{Experimental Setup}
This subsection describes the datasets, baseline models, evaluation metrics, and implementation details used in the experiments.

\subsubsection{Datasets}
 We conduct experiments on Weibo-21, Fakeddit, and SSS. Weibo-21 \citep{nan2021mdfend} and Fakeddit \citep{nakamura2020fakeddit} are Chinese and English multimodal fake news datasets, respectively. Because some samples were published long ago and contain invalid links or damaged content, we apply a unified cleaning procedure to these datasets. SSS is a Chinese--English multimodal fake news detection dataset constructed from six sources, including Weibo-17 \citep{jin2017multimodal}, Weibo-21, CFND \citep{zhang2024natural}, MR2 \citep{hu2023mr2}, Fakeddit, and FineFake \citep{zhou2026finefake}, through sample deduplication, quality screening, and manual review. Details of its construction are provided in Appendix~A. Harm labels for all three datasets were systematically annotated by human annotators; details are provided in Appendix~B. Table~\ref{tab:dataset-scale} reports the numbers of news and evidence items used in the experiments.

\begin{table}[htbp]
\centering
\small
\caption{Numbers of news and automatically retained evidence on the three datasets.}
\label{tab:dataset-scale}
\renewcommand{\arraystretch}{1.2}
\begin{tabular}{lccc}
\toprule
Components & Weibo-21 & Fakeddit & SSS \\
\midrule
News     & 4,493 & 5,919  & 7,997 \\
Evidence & 8,341 & 11,375 & 15,157 \\
\bottomrule
\end{tabular}
\end{table}

\begin{table*}[t]
\centering
\small
\caption{Overall performance comparison of RAEGNet and baselines on the three datasets. ``w/o EK'' denotes methods that do not use external knowledge, ``w/ EK'' denotes methods that incorporate external knowledge, and ``w/ LLM'' denotes LLM-based methods. RAEGNet-E, RAEGNet-B, and RAEGNet-A use entity-level evidence, pre-news event-level evidence, and full event-level evidence, respectively.}
\label{tab:overall_performance}

\begin{tabular}{cl *{9}{c}}
\toprule
\multirow{3}{*}{Category}
& \multirow{3}{*}{Method}
& \multicolumn{3}{c}{Weibo-21}
& \multicolumn{3}{c}{Fakeddit}
& \multicolumn{3}{c}{SSS} \\

\cmidrule(lr){3-5}
\cmidrule(lr){6-8}
\cmidrule(lr){9-11}

& & \multirow{2}{*}{Accuracy}
& \multicolumn{2}{c}{F1-score}
& \multirow{2}{*}{Accuracy}
& \multicolumn{2}{c}{F1-score}
& \multirow{2}{*}{Accuracy}
& \multicolumn{2}{c}{F1-score} \\

\cmidrule(lr){4-5}
\cmidrule(lr){7-8}
\cmidrule(lr){10-11}

& & & Fake & Real
& & Fake & Real
& & Fake & Real \\
\midrule
\multirow{4}{*}{w/o EK}
& CAFE
& 0.815 & 0.810 & 0.820
& 0.786 & 0.816 & 0.745
& 0.726 & 0.753 & 0.693 \\

& MRML
& 0.903 & 0.908 & 0.898
& 0.860 & 0.876 & 0.839
& 0.758 & 0.779 & 0.732 \\

& Event-Radar
& 0.881 & 0.884 & 0.877
& 0.840 & 0.859 & 0.815
& 0.754 & 0.767 & 0.739 \\

& MSACA
& 0.894 & 0.900 & 0.886
& 0.846 & 0.864 & 0.823
& 0.762 & 0.778 & 0.744 \\

\midrule

\multirow{3}{*}{w/ EK}
& KEHGNN-FD
& 0.764 & 0.776 & 0.751
& 0.763 & 0.796 & 0.716
& 0.674 & 0.698 & 0.644 \\

& NSLM
& 0.878 & 0.885 & 0.870
& 0.850 & 0.867 & 0.828
& 0.781 & 0.804 & 0.750 \\

& ERIC-FND
& 0.844 & 0.857 & 0.828
& 0.794 & 0.828 & 0.744
& 0.655 & 0.724 & 0.539 \\

\midrule

\multirow{3}{*}{w/ LLM}
& Qwen2.5-VL
& 0.742 & 0.754 & 0.729
& 0.738 & 0.771 & 0.694
& 0.629 & 0.664 & 0.587 \\

& InternVL2.5
& 0.714 & 0.726 & 0.701
& 0.726 & 0.763 & 0.676
& 0.613 & 0.650 & 0.568 \\

& GLPN-LLM
& 0.882 & 0.888 & 0.874
& 0.875 & 0.888 & 0.858
& 0.707 & 0.743 & 0.660 \\

\midrule

\multirow{3}{*}{Ours}
& RAEGNet-E
& 0.911 & 0.911 & 0.911
& 0.898 & 0.908 & 0.886
& 0.808 & 0.817 & 0.798 \\

& RAEGNet-B
& 0.928 & 0.930 & 0.926
& 0.911 & 0.922 & 0.898
& 0.837 & 0.851 & 0.819 \\

& RAEGNet-A
& 0.931 & 0.934 & 0.929
& 0.919 & 0.926 & 0.910
& 0.843 & 0.860 & 0.822 \\

\bottomrule
\end{tabular}
\end{table*}

\subsubsection{Baselines}
We select several task-specific methods as comparison baselines. Methods that do not use external knowledge include CAFE \citep{chen2022cross}, MRML \citep{peng2023mrml}, Event-Radar \citep{ma2024event}, and MSACA \citep{wang2024fake}. Methods that use external knowledge include KEHGNN-FD \citep{xie2023knowledge}, NSLM \citep{dong2024unveiling}, and ERIC-FND \citep{cao2025external}.

In addition, we include three LLM-based baselines: Qwen2.5-VL-7B-Instruct \citep{bai2025qwen}, InternVL2.5-8B \citep{chen2024expanding}, and GLPN-LLM \citep{hu2025synergizing}. Qwen2.5-VL-7B-Instruct and InternVL2.5-8B are locally deployable open-source multimodal LLMs that directly judge news authenticity from the news text, news image, and retrieved evidence, without task-specific fine-tuning. GLPN-LLM is included in the same category as an LLM-based external-reasoning baseline.

Each method is evaluated separately on each dataset using the same five-fold cross-validation partitions, and its parameters follow either the settings in the corresponding paper or those in the public implementation. External-knowledge methods use their original external-information acquisition mechanisms as specified in their papers, whereas RAEGNet uses the event-level evidence retrieval method proposed in this work. Direct LLM baselines are not fine-tuned; they use the same retrieved evidence cache and a fixed judgment prompt for the held-out samples in each fold.

\subsubsection{Evaluation metrics}
We use the conventional metrics of accuracy, precision, recall, and F1 score. For harm-level evaluation, we define the high-harm subset as all held-out samples with $h_i\geq0.7$ in each fold. Within this subset, HHF Precision, HHF Recall, and HHF F1 are computed by treating fake news as the positive class and high-harm real news as the negative class.

\begin{table*}[t]
\centering
\small
\caption{Ablation study on the Weibo-21, Fakeddit, and SSS datasets. Indented rows prefixed by ``--'' indicate that only the specified subcomponent of the corresponding higher-level module is ablated.}
\label{tab:ablation_study}

{
\begin{tabular}{l *{9}{c}}
\toprule

\multirow{3}{*}{Ablation Settings}
& \multicolumn{3}{c}{Weibo-21}
& \multicolumn{3}{c}{Fakeddit}
& \multicolumn{3}{c}{SSS} \\

\cmidrule(lr){2-4}
\cmidrule(lr){5-7}
\cmidrule(lr){8-10}

& \multirow{2}{*}{Accuracy}
& \multicolumn{2}{c}{HHF}
& \multirow{2}{*}{Accuracy}
& \multicolumn{2}{c}{HHF}
& \multirow{2}{*}{Accuracy}
& \multicolumn{2}{c}{HHF} \\

\cmidrule(lr){3-4}
\cmidrule(lr){6-7}
\cmidrule(lr){9-10}

& & F1 & Rec.
& & F1 & Rec.
& & F1 & Rec. \\
\midrule
w/o Evidence Graph
& 0.913 & 0.956 & 0.961
& 0.909 & 0.939 & 0.952
& 0.822 & 0.892 & 0.956 \\

\quad -- w/o Edge Weights
& 0.920 & 0.959 & 0.956
& 0.915 & 0.936 & 0.935
& 0.837 & 0.886 & 0.923 \\

\quad -- w/o Edge Types
& 0.914 & 0.960 & 0.967
& 0.911 & 0.930 & 0.935
& 0.826 & 0.877 & 0.928 \\

\midrule

w/o Harm-aware Branch
& 0.915 & 0.946 & 0.919
& 0.912 & 0.925 & 0.907
& 0.839 & 0.878 & 0.905 \\

\quad -- w/o $\mathcal{L}_{\mathrm{harm}}$
& 0.923 & 0.954 & 0.936
& 0.909 & 0.926 & 0.919
& 0.839 & 0.888 & 0.923 \\

\quad -- w/o $\mathcal{L}_{\mathrm{rank}}$
& 0.924 & 0.947 & 0.932
& 0.906 & 0.922 & 0.900
& 0.833 & 0.892 & 0.917 \\

\midrule

w/o News Text
& 0.886 & 0.919 & 0.891
& 0.870 & 0.894 & 0.935
& 0.815 & 0.891 & 0.910 \\

w/o News Image
& 0.859 & 0.925 & 0.919
& 0.860 & 0.894 & 0.943
& 0.803 & 0.883 & 0.927 \\

w/o Evidence
& 0.907 & 0.954 & 0.933
& 0.886 & 0.929 & 0.942
& 0.816 & 0.888 & 0.904 \\

\midrule

Full Model
& 0.931 & 0.972 & 0.974
& 0.919 & 0.947 & 0.955
& 0.843 & 0.896 & 0.952 \\

\bottomrule
\end{tabular}
}
\end{table*}

\begin{figure*}[t]
    \centering
    \includegraphics[width=\textwidth]{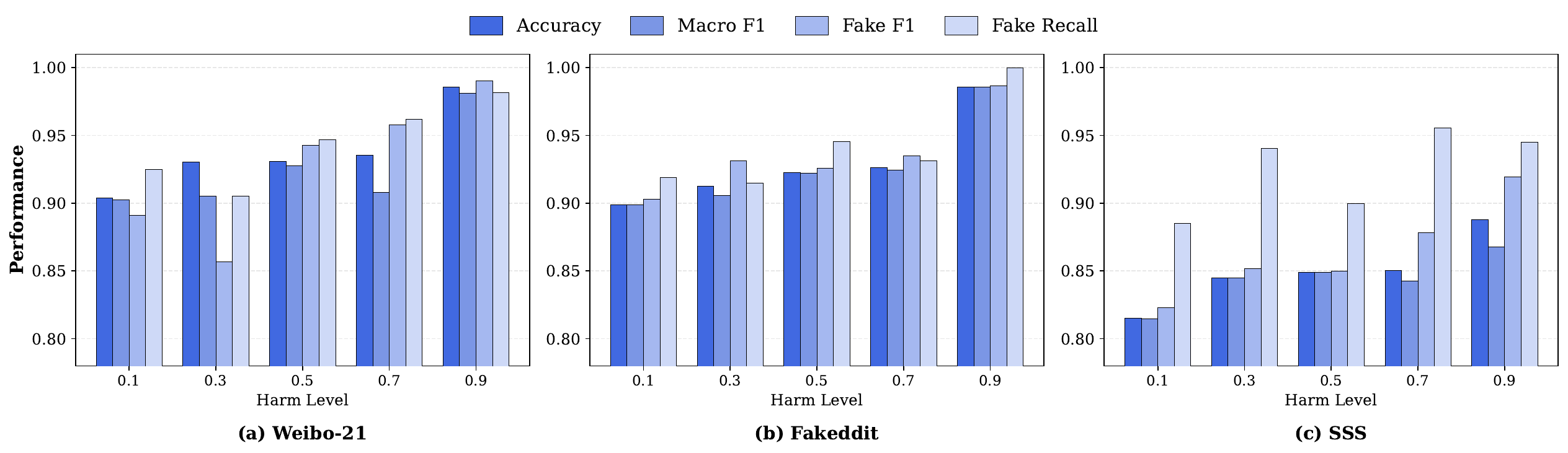}
    \caption{Detection performance of RAEGNet at different harm levels.}
    \label{fig:harm-level-performance}
\end{figure*}

\subsubsection{Implementation details}
We use CLIP-ViT-L/14 \cite{radford2021learning} and Chinese-CLIP-ViT-L/14 \cite{yang2022chinese} to encode English and Chinese samples, respectively. All task-specific baselines, RAEGNet variants, and ablation models are evaluated using the same five-fold cross-validation partitions, and the results are reported as the mean and standard deviation over the five held-out folds. Direct open-source MLLM baselines are evaluated in an evidence-augmented zero-shot setting on the same held-out folds.

RAEGNet-E uses entity-level evidence, RAEGNet-B uses pre-news event-level evidence and serves as the main early-detection setting, while RAEGNet-A uses both pre-news and post-news event-level evidence only for post-hoc fact-checking. The two event-level variants correspond to different information-availability scenarios. Detailed model hyperparameters, evidence construction procedures, and MLLM prompts are provided in Appendices~\ref{sec:event_level_retrieval} and \ref{sec:detailed_experimental_settings}.

\subsection{Overall Performance}
Table~\ref{tab:overall_performance} presents the overall detection results of RAEGNet and the baseline methods on the three datasets. RAEGNet-B consistently outperforms the baselines in the early-detection setting, showing that event-level evidence remains useful when only evidence published before the corresponding news item is available. RAEGNet-A achieves the strongest performance in the post-hoc fact-checking setting, where a broader set of filtered evidence can be used for evidence-supported verification. Compared with RAEGNet-E, which uses entity-level evidence under the same model architecture, both event-level variants achieve higher accuracy and F1 scores on all three datasets.

\subsection{Ablation Study}
We construct nine ablation variants. The results show that the news text, image, and external evidence all make substantial contributions to final performance. Removing the evidence graph, edge types, or edge weights generally reduces accuracy and HHF F1. This finding indicates that the relation-aware evidence graph and its edge-feature design improve authenticity detection. By contrast, removing the harm-aware branch, the harm-calibration loss, or the ordinal ranking loss reduces HHF F1 and HHF Recall, indicating that conditional-harm modeling provides useful supervision for high-harm fake-news recognition.

\subsection{Harm-Level Analysis}
Figure~\ref{fig:harm-level-performance} presents the detailed performance of RAEGNet at different harm levels. A higher level indicates greater potential harm if the content is false and disseminated. The results show that RAEGNet achieves higher detection accuracy and better overall recognition performance for fake news with greater potential harm. In addition to authenticity classification, the harm-risk branch and ordinal ranking loss explicitly model the magnitude and ordinal structure of conditional harm, providing complementary supervision for high-harm fake-news recognition.

\subsection{Effectiveness of Event-Level Retrieval}
\label{sec:event_retrieval_effectiveness}
To evaluate the effectiveness of event-level retrieval, we compare entity-level and event-level evidence under the same RAEGNet architecture, keeping the model architecture and training configuration unchanged. The entity-level setting uses evidence retrieved by ERIC-FND, whereas the event-level setting uses leakage-filtered evidence retrieved by ELERF. As shown in Table~\ref{tab:event_vs_entity_retrieval}, event-level retrieval consistently improves accuracy and fake/real F1 scores across all three datasets, indicating that complete-event queries retrieve more relevant evidence and reduce entity-related but event-irrelevant noise.

\begin{table}[t]
\centering
\small
\caption{Comparison between entity-level and event-level evidence retrieval under the same RAEGNet architecture.}
\label{tab:event_vs_entity_retrieval}
\begin{tabular}{llccc}
\toprule
Dataset & Retrieval & Acc. & Fake F1 & Real F1 \\
\midrule
\multirow{2}{*}{Weibo-21}
& Entity-level & 0.911 & 0.911 & 0.911 \\
& Event-level  & 0.931 & 0.934 & 0.929 \\
\midrule
\multirow{2}{*}{Fakeddit}
& Entity-level & 0.898 & 0.908 & 0.886 \\
& Event-level  & 0.919 & 0.926 & 0.910 \\
\midrule
\multirow{2}{*}{SSS}
& Entity-level & 0.808 & 0.817 & 0.798 \\
& Event-level  & 0.843 & 0.860 & 0.822 \\
\bottomrule
\end{tabular}
\end{table}

\begin{figure}[t]
    \centering
    \includegraphics[width=\columnwidth]{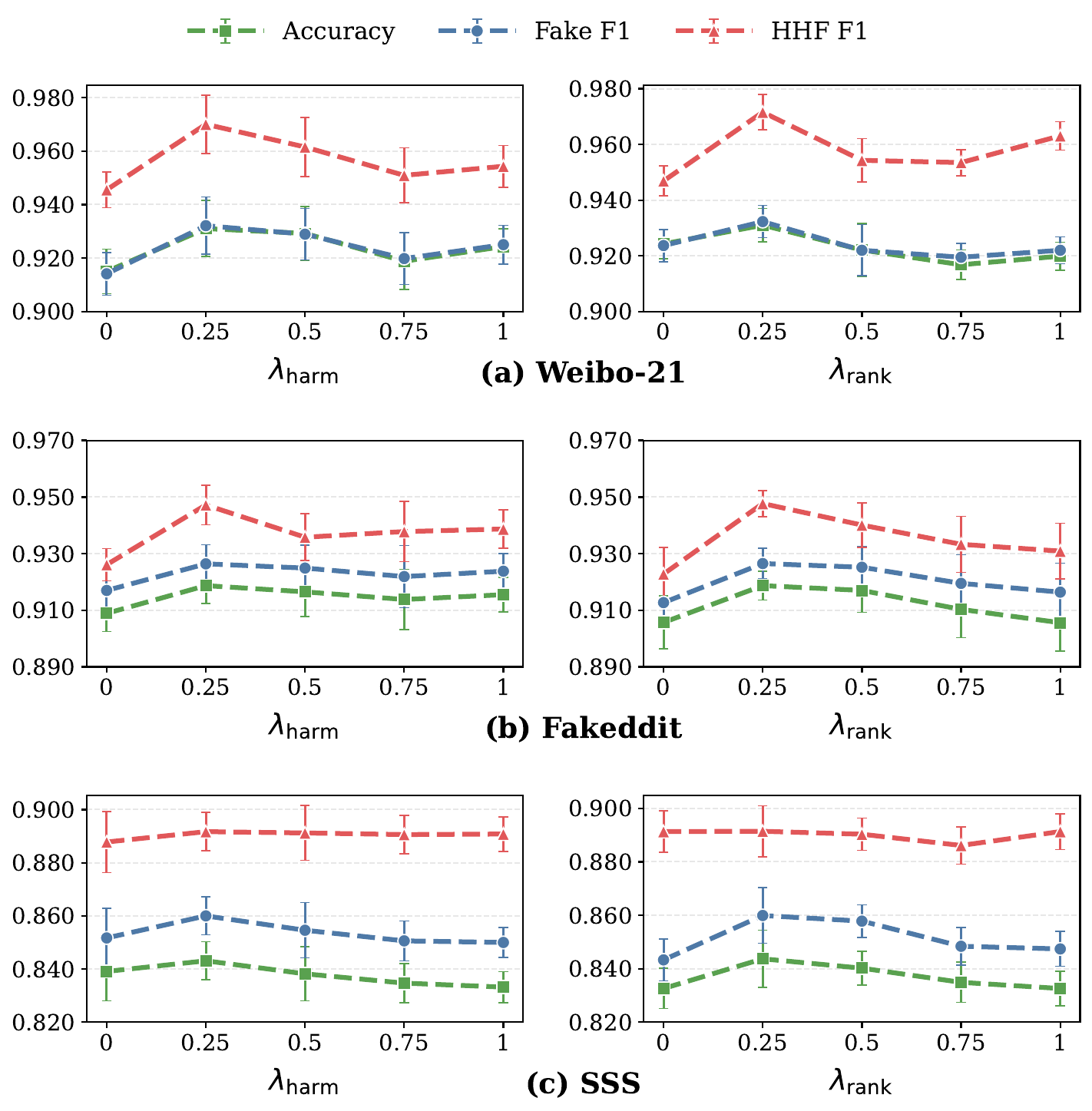}
    \caption{Sensitivity analysis of $\lambda_{\mathrm{harm}}$ and $\lambda_{\mathrm{rank}}$.}
    \label{fig:sensitivity}
\end{figure}

\begin{figure}[t]
    \centering
    \includegraphics[width=\columnwidth]{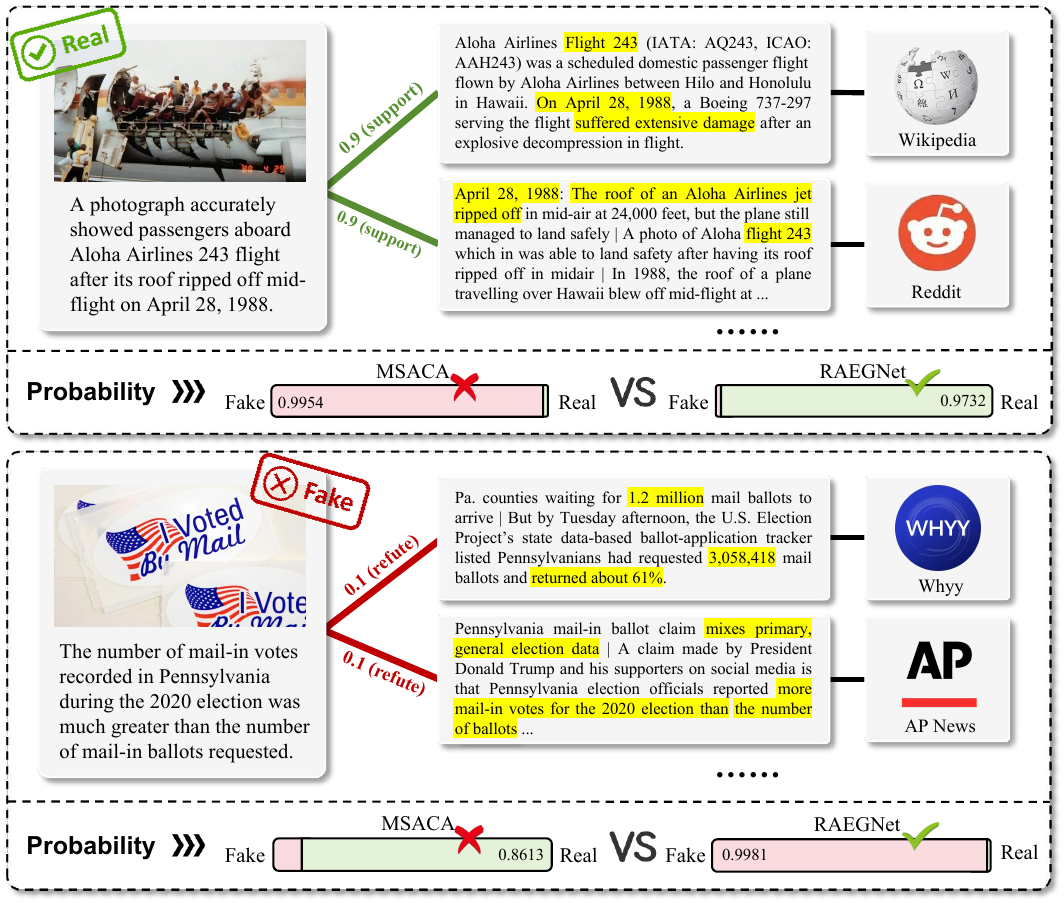}
    \caption{Case study.}
    \label{fig:case_study}
\end{figure}

\subsection{Sensitivity Analysis}
We further analyze the effects of the harm-calibration loss weight $\lambda_{\mathrm{harm}}$ and the ordinal-ranking loss weight $\lambda_{\mathrm{rank}}$. As shown in Figure~\ref{fig:sensitivity}, $\lambda_{\mathrm{harm}}=0.25$ and $\lambda_{\mathrm{rank}}=0.25$ yield relatively balanced results across all three datasets. Weights that are too small weaken conditional-harm supervision, whereas excessively large weights may impair the model's basic authenticity classification. The consistent trends across the three datasets indicate that the selected weights provide a stable balance between authenticity classification and conditional-harm modeling.

\subsection{Case Study}
As shown in Figure~\ref{fig:case_study}, the upper example is a real news report concerning Aloha Airlines Flight 243, for which the news item and the retrieved evidence are highly consistent in stance. With the assistance of external evidence, RAEGNet correctly classifies the news as real, whereas MSACA produces an incorrect prediction. The lower example is a false claim concerning mail-in ballots in the 2020 U.S. election. RAEGNet makes the correct judgment using information from the retrieved evidence, whereas MSACA makes an incorrect prediction.

\section{Conclusion}
We propose RAEGNet, which unifies a relation-aware evidence graph with conditional-harm learning for fake news detection. The model enhances information interaction between news and evidence through multiple relation types and edge weights, while jointly modeling harm-risk assessment and authenticity classification. Experiments on three datasets demonstrate that RAEGNet outperforms the baselines overall and exhibits strong capability in identifying high-harm fake news.

\section*{Limitations}
The model relies on the quality of external retrieval and LLMs. Our event-level evidence acquisition depends on the initial recall of search engines, and the quantification of stance scores requires the assistance of LLMs. Additionally, the introduction of LLMs inevitably increases the inference latency of the system, which poses computational efficiency challenges for real-time streaming data detection scenarios with high concurrency.

\section*{Ethical Considerations}
This study uses publicly available research datasets and web-accessible evidence solely for academic research on misinformation detection. Data collection and use followed the access conditions, licenses, and privacy requirements of the original datasets and platforms, and no private user attributes were inferred. The proposed framework is intended exclusively for defensive misinformation analysis and should not be used for surveillance, censorship, or automated decision-making without appropriate human oversight.

\bibliography{anthology,custom}
\bibliographystyle{acl_natbib}
\clearpage

\appendix

\section{Construction of the SSS Dataset}
During this study, we found that existing multimodal fake news detection datasets contain a large amount of low-quality data, including personal emotional expressions, fragmented statements lacking a verifiable factual subject, and damaged samples with missing key modalities. Such low-quality data cannot faithfully and effectively reflect model capability. To better evaluate practical performance, we screened and assembled a high-quality SSS (Strictly Selected Subset) dataset from several existing public datasets. The construction procedure is described below.

\subsection{Basic Data Cleaning}
In the initial phase, we collected six open-source datasets, including Fakeddit, FineFake, MR2, Weibo-17, Weibo-21, and CFND, with an initial total scale of over 1.14 million items. To remove obvious noise at an early stage, we designed the following basic cleaning rules:

\noindent Textual Part:
\begin{itemize}
    \item Regular expressions were used to precisely remove residual HTML tags, special invisible characters, and garbled text.
    \item Texts shorter than 15 words or Chinese characters were removed to ensure that each sample contained a basic semantic context.
\end{itemize}
\noindent Visual Part:
\begin{itemize}
    \item Samples whose original image links were invalid, whose image files were corrupted, or whose images could not be decoded normally were automatically detected and removed.
\end{itemize}

\noindent Multimodal Deduplication:
\begin{itemize}
    \item To reduce near-duplicate leakage while avoiding the removal of meaningful variations, we encoded each textual claim and its associated image into a shared multimodal representation and computed cosine similarity between candidate sample pairs. Pairs with multimodal similarity higher than 0.85 were selected as potential near-duplicates and manually reviewed. A pair was removed as a duplicate only when the two samples expressed the same factual claim and contained the same or nearly identical visual context.
\end{itemize}

\begin{table}[!t]
\centering
\footnotesize
\caption{The structured system prompt and examples for LLM-based semantic filtering.}
\label{tab:llm_prompt}
\begin{tabular}{c p{0.75\columnwidth}}
\toprule
Role & Content \\
\midrule
\multirow[c]{35}{*}{System} & You are an expert in fake news research data screening, dedicated to filtering high-quality multimodal news data.\\
& Each input contains a news text and its associated news image. Evaluate the two modalities jointly.\\
& \\
& I. Judgment Criteria:\\
& 1. Applicable conditions (must be fully met to return ``Usable''):\\
& (1) The text must possess a clear news format and complete structure. The selected content cannot be fragmented isolated words, but must have a clear description of the news event, and contain the basic elements constituting the news.\\
& (2) The content needs to present a complete chain of factual statements. Qualified data cannot be merely isolated subjective assertions or emotional venting.\\
& 2. Exclusion conditions (if any is met, immediately return ``Unusable''):\\
& (1) The text is incomplete, missing, or truncated.\\
& (2) There are obvious grammatical errors or logical confusion.\\
& (3) Lacks basic news elements.\\
& (4) Lacks specific details, making truth verification impossible.\\
& (5) Contains a large amount of garbled text or special characters.\\
& \\
& II. Return Requirements:\\
& 1. Only return ``Usable'' or ``Unusable'', no need to explain the reason.\\
& 2. If it cannot be judged, return ``-1''. \\
\midrule
User & News text: ``A strong 7.8 magnitude earthquake occurred in southern Turkey and northern Syria early Monday morning. Multiple buildings collapsed, and rescue teams are being deployed.'' \newline News image: \texttt{[NEWS IMAGE]} \\
\midrule
LLM & Usable \\
\midrule
User & News text: ``It rained today, and I feel very sad. I hate this weather, it makes me want to stay in bed all day.'' \newline News image: \texttt{[NEWS IMAGE]} \\
\midrule
LLM & Unusable \\
\midrule
User & News text: ``A new policy was released yesterday.'' \newline News image: \texttt{[NEWS IMAGE]} \\
\midrule
LLM & Unusable \\
\bottomrule
\end{tabular}
\end{table}

\begin{table*}[t]
\centering
\small
\caption{Specific distribution and volume changes of each data source before and after screening.}
\label{tab:screening_statistics}
\begin{tabular}{lcccc}
\toprule
Dataset & Language & Original Quantity & Filtered Quantity\\
\midrule
Fakeddit \citep{nakamura2020fakeddit} & English & 1,063,106 & 2,922 \\
FineFake \citep{zhou2026finefake} & English & 16,909 & 2,461 \\
MR2 \citep{hu2023mr2} & Chinese \& English & 14,700 & 442 \\
Weibo-17 \citep{jin2017multimodal} & Chinese & 9,528 & 942 \\
Weibo-21 \citep{nan2021mdfend} & Chinese & 9,128 & 850 \\
CFND \citep{zhang2024natural} & Chinese & 26,665 & 380 \\
\midrule
Total & Chinese \& English & 1,140,036 & 7,997 \\
\bottomrule
\end{tabular}
\end{table*}

\begin{table*}[t]
\centering
\small
\caption{Statistics of similarity-assisted event grouping used for fold construction. Candidate pairs are retrieved automatically, whereas event identity is determined by human reviewers.}
\label{tab:event_audit_stats}
\begin{tabular}{lrrrr}
\toprule
Dataset & Samples & Candidate Pairs & Verified Event Groups & Grouped Samples \\
\midrule
Weibo-21 & 4,493 & 286 & 74 & 198 \\
Fakeddit & 5,919 & 413 & 96 & 271 \\
SSS & 7,997 & 672 & 143 & 406 \\
\bottomrule
\end{tabular}
\end{table*}

\subsection{LLM-Based Semantic Filtering}
To ensure that the samples possessed a complete news structure and fact-checking value, we introduced a multimodal large language model as a data-screening expert for in-depth semantic filtering. The model jointly evaluates the news text and its associated image. The screening criteria were converted into a structured system prompt. The detailed criteria and examples are shown in Table~\ref{tab:llm_prompt}.

\subsection{Manual Review}
After semantic filtering by the large language model, we organized five reviewers with research backgrounds in misinformation detection to independently assess the samples further, thereby ensuring the reliability of the final retained data. Manual review focused on re-examining the model's decisions and reaching final judgments on uncertain samples. At the decision stage, a sample was retained only if it received explicit approval from at least four reviewers. Samples with disagreement were discussed collectively by the group, which then made the final decision. Ultimately, 7,893 usable samples were selected from the 8,597 samples judged usable by the LLM, and 104 usable samples were selected from the 312 uncertain samples.

\subsection{Event Grouping and Fold Construction}
After manual review, we performed a semantic-similarity-assisted event grouping procedure before constructing the five cross-validation folds. For each news item, we retrieve its nearest neighbors according to multimodal cosine similarity and retain pairs with similarity higher than 0.85 as candidate event-related pairs. This threshold is used only to improve the recall of potentially related samples, rather than to automatically merge them. Human reviewers then determine whether each candidate pair describes the same real-world event by considering the core entities, main event action, time, location, and factual content. Reports that are rewritings, shortened versions, or multimodal variants of the same event are assigned to the same event group, whereas samples that only share a broad topic or similar wording are kept as distinct events.

During fold construction, all samples assigned to the same event group are placed in the same fold. The allocation across the five folds preserves the overall distributions of authenticity labels, languages, source datasets, and harm levels. Table~\ref{tab:event_audit_stats} reports the number of candidate pairs inspected during this procedure. The grouping procedure substantially reduces direct event overlap between training and held-out folds.

\subsection{Analysis of Dataset Composition}
After the three stages described above, the final SSS (Strictly Selected Subset) dataset contained 7,997 high-quality multimodal news samples, including 3,836 real and 4,161 fake samples. In terms of language, 2,614 samples are in Chinese and 5,383 are in English, and the average harm coefficient of the entire dataset is 0.489. The relatively balanced authenticity distribution mitigates severe class imbalance, while the bilingual composition supports evaluation across different linguistic contexts. Table~\ref{tab:screening_statistics} reports the original and retained numbers of samples from each data source.

\begin{table*}[t]
\centering
\small
\caption{Annotation criteria for harm levels.}
\label{tab:harm-guidelines}
\begin{tabularx}{\textwidth}{c l X}
\toprule
Harm label & Level & Typical content \\
\midrule
0.1 & Very low harm & Entertainment gossip, minor misleading content, and everyday misinformation with a limited scope of influence. \\
\midrule
0.3 & Low harm & Content that may cause localized misunderstanding, but the impact is limited and the harm is controllable. \\
\midrule
0.5 & Moderate harm & Content involving public events, social controversies, or collective perceptions that may provoke disputes or cause a certain degree of adverse impact. \\
\midrule
0.7 & High harm & Content involving public safety, health, politics, disasters, or intergroup conflict that may produce a clear social impact. \\
\midrule
0.9 & Very high harm & Content that may induce panic, real-world harm, risks to public order, or large-scale collective impact. \\
\bottomrule
\end{tabularx}
\end{table*}

\begin{table*}[t]
\centering
\small
\caption{Agreement and adjudication rates for harm labels.}
\label{tab:harm-agreement}
\begin{tabular}{lccc}
\toprule
Evaluation & Weibo-21 & Fakeddit & SSS \\
\midrule
Human annotator agreement (Krippendorff's $\alpha$)
  & 0.9598 & 0.9303 & 0.9413 \\
\addlinespace[2pt]
LLM--human agreement (mean agreement)
  & 92.02\% & 89.37\% & 90.13\% \\
\addlinespace[2pt]
Adjudication Trigger Rate
  & 0.78\% (35/4,493) & 1.32\% (78/5,919) & 1.25\% (100/7,997) \\
\bottomrule
\end{tabular}
\end{table*}

\section{Quantification of Harm Coefficients}
\subsection{Definition of Harm Labels}
In real-world online ecosystems, different instances of fake news vary substantially in their potential destructive impact. For example, political rumors and public-health misinformation are far more harmful than entertainment gossip. To characterize this difference, we define \emph{conditional harm} as the potential degree of social impact that the content could cause if it were false and disseminated through social networks, without presupposing the current authenticity of the news item.

\subsection{Manual Annotation}
Harm labels were independently assigned by five annotators with research backgrounds in misinformation detection. Under a unified standard, the annotators assigned every news item to one of five ordered levels, which were mapped to $h\in\{0.1,0.3,0.5,0.7,0.9\}$. Under normal circumstances, the mean of the five annotators' level assignments was first computed and then mapped to the nearest discrete harm level as the final label of the sample. If the difference between the highest and lowest levels assigned by the five annotators was two levels or more, the sample was considered to exhibit substantial disagreement. The five annotators then discussed the sample as a group and jointly determined its final label.

\subsection{Automated LLM-Based Annotation}
In addition to manual labels, we design an automated news-harm annotation procedure based on a large language model. Its purpose is to reduce the cost of manual annotation when transferring the method to other datasets and to improve adaptability to newly added data. The automated procedure uses the same definition of conditional harm and asks the model to output a continuous harm score in $[0,100]$ under the assumption that the news is entirely false and is widely disseminated. The continuous score is then normalized and mapped to $\widetilde h\in\{0.1,0.3,0.5,0.7,0.9\}$ to obtain the automated harm label. It should be emphasized that the final supervision labels used in our experiments are the labels produced by the five human annotators. The LLM-based procedure is used mainly to validate the feasibility of automated scaling and serves as an auxiliary tool for rapidly obtaining initial harm labels on larger datasets in future work.

\subsection{Annotation Agreement and Validation of the Automated Procedure}
To evaluate the reliability of the harm labels, we measure inter-annotator agreement, agreement between the automated LLM labels and human labels, and the proportion of samples requiring adjudication. Inter-annotator agreement is measured using Krippendorff's alpha. LLM-human agreement denotes the agreement between the automated LLM labels and the mean labels of the five human annotators. The Adjudication Trigger Rate denotes the number and percentage of samples for which group adjudication was triggered because the difference between the maximum and minimum levels assigned by the five annotators was two levels or more.

The results show that inter-annotator agreement satisfies $\alpha>0.90$ on all three datasets, indicating that the definitions of the harm levels are highly operational. The proportion of samples that trigger group adjudication is below 1.5\% on every dataset, showing that samples with substantial disagreement are rare. Mean agreement between the automated LLM labels and human labels is close to or above 90\%, suggesting that the automated procedure can approximate human risk judgments well.

\subsection{Distribution of Harm Coefficients}
\begin{figure}[htbp]
    \centering
    \includegraphics[width=\columnwidth]{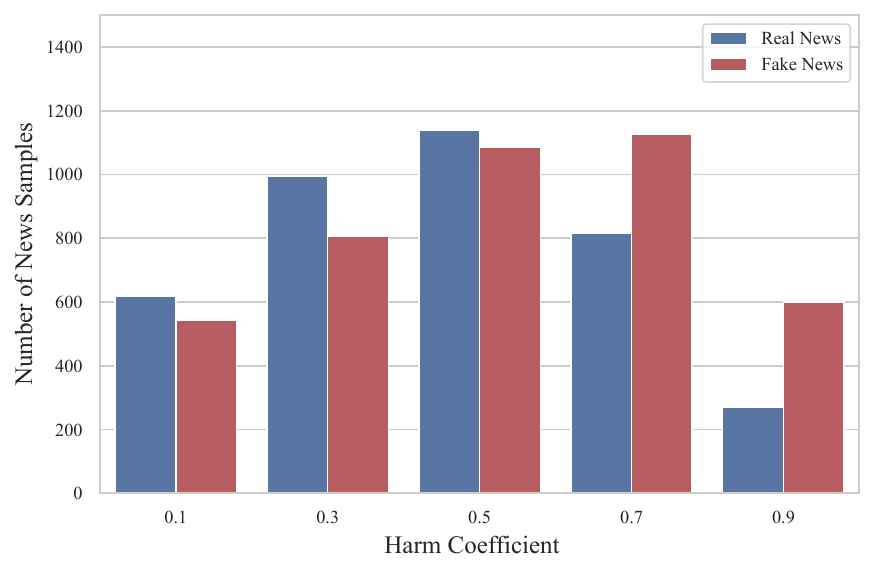}
    \caption{Distribution of harm coefficients in the SSS dataset.}
    \label{fig:harm_distribution}
\end{figure}

To visualize the distribution of news harmfulness in the SSS dataset, we count the numbers of real- and fake-news samples at each discrete harm coefficient. The results are shown in Figure~\ref{fig:harm_distribution}. Real-news samples are concentrated mainly in the range from 0.3 to 0.7, whereas the number of fake-news samples increases markedly at the high-harm levels of 0.7 and 0.9.

This distribution reflects the empirical risk structure observed in the datasets used in this study. In realistic information environments, false claims may naturally concentrate in high-impact topics such as public health, disasters, politics, and social conflicts. Conditional harm is therefore treated as an intrinsic risk attribute of the news content under the assumption that the content is false and widely disseminated, rather than as a variable that must be artificially decorrelated from authenticity.

At the same time, we examine whether the model benefits only from the marginal association between harm levels and authenticity labels. For this purpose, we conduct harm-matched evaluation, where real and fake samples are balanced within each harm level in the held-out set. The full model remains stronger than the variant without Harm-aware Branch under this matched setting, suggesting that conditional-harm supervision contributes beyond the marginal association between harm and authenticity labels.

\section{Event-Level Evidence Retrieval Framework}
\label{sec:event_level_retrieval}

\subsection{Retrieval}
We use Brave Search as the underlying search engine and restrict retrieval to a predefined source pool fixed before evaluation. The source pool covers encyclopedic resources, news media, official institutions, academic sources, specialized domains, and public communication platforms, as illustrated in Table~\ref{tab:retrieval_domains}. This design avoids post-hoc source selection for individual held-out samples and makes the retrieval scope auditable.

For each news item, we use the complete news text as the query and retain the top 10 relevant webpage snippets together with their source and temporal metadata. We first remove the target news URL when available, same-domain reposts or syndicated copies, and near-duplicate candidates whose text or visual content is highly similar to the target news item. The remaining candidates are then partitioned according to their temporal relation to the corresponding news item. Evidence published before the news item is assigned to the pre-news evidence set, whereas evidence published after the news item is assigned to the post-news evidence set. This temporal partition is performed before leakage filtering, and both subsets subsequently undergo the same automatic explicit filtering, LLM-based implicit leakage screening, and stance quantification procedures.

Evidence whose publication time cannot be reliably determined is excluded from the pre-news evidence set and is therefore not used by RAEGNet-B. Such evidence may be considered for RAEGNet-A only after passing the complete leakage-control procedure.

\begin{table*}[!t]
\centering

\small
\caption{Retrieval domains organized by primary and secondary categories.}
\label{tab:retrieval_domains}

\renewcommand{\arraystretch}{1.25}
\renewcommand{\tabularxcolumn}[1]{m{#1}}

\begin{tabularx}{\textwidth}{
    |>{\centering\arraybackslash}m{0.14\textwidth}
    |>{\centering\arraybackslash}m{0.14\textwidth}
    |>{\raggedright\arraybackslash}X|
}
\hline

\textbf{Primary Category}
& \textbf{Secondary Category}
& \centering\arraybackslash\textbf{Retrieval Domains} \\
\hline

\multicolumn{2}{|c|}{Encyclopedias and Knowledge Bases}
& wikipedia, britannica, baidu \\
\hline

\multicolumn{2}{|c|}{Communities and Content Platforms}
& zhihu, douban, reddit, quora, youtube, weibo, sina, tencent, 163, sohu \\
\hline

\multirow[c]{11}{0.14\textwidth}
{\centering News Media}
& News Agencies
& reuters, apnews, xinhuanet, chinanews, tass \\
\cline{2-3}

& Transnational Media
& bbc, cnn, aljazeera, dw, euronews, france24, rt, sputniknews,
  chinadaily, sky, ifeng \\
\cline{2-3}

& National Media
& cctv, people, thepaper, huanqiu, guancha, gmw, nytimes,
  washingtonpost, theguardian, foxnews, cbsnews, nbcnews, npr,
  pbs, time, newsweek, independent, dailymail, usatoday,
  telegraph, express, mirror, huffpost, cbc, theglobeandmail,
  ndtv, thehindu, hindustantimes, indianexpress, indiatoday,
  straitstimes, japantimes, haaretz, timesofisrael, jpost,
  elpais, nzherald, buzzfeednews, vice, vox, slate, theatlantic,
  buzzfeed \\
\cline{2-3}

& Local and Regional Media
& bjnews, oeeee, scmp, chicagotribune, boston, bostonglobe,
  miamiherald, syracuse, denverpost, seattletimes, dallasnews,
  houstonchronicle, tampabay, kansascity, cleveland, oregonlive,
  sltrib, latimes, bangordailynews, nola, jacksonville,
  mercurynews, thestar, torontosun, vancouversun, smh, abc7,
  abc7ny, ktla, cbslocal, abc13, 9news, whyy \\
\hline

\multirow[c]{9}{0.14\textwidth}
{\centering Specialized Domains}
& Government and Military
& whitehouse, fbi, justice, state, senate, house, army, navy,
  congress, thehill, politico \\
\cline{2-3}

& Business and Finance
& jiemian, ce, nbd, yicai, eastmoney, businessinsider, cnbc,
  forbes, fortune, bloomberg, qz, wsj \\
\cline{2-3}

& Entertainment
& rollingstone, variety, hollywoodreporter \\
\cline{2-3}

& Technology
& theverge, wired, techcrunch, mashable, gizmodo, arstechnica,
  cnet \\
\cline{2-3}

& Academic Sources
& nature, science, pewresearch, brookings, hoover, harvard,
  yale, stanford, cornell, mit, ucsb, umich \\
\cline{2-3}

& Others
& nationalgeographic, smithsonianmag, history,
  allthatsinteresting, motherjones, loc, archives, amnesty,
  hrw, aclu, nationalww2museum, ushmm, nih, cdc, nasa, cma,
  guokr, kepuchina, livescience, iflscience,
  scientificamerican \\
\hline

\end{tabularx}

\vspace{0.2em}

\caption{Full prompt used for LLM-based implicit leakage screening.}
\label{tab:implicit_leakage_prompt}
\begin{tabularx}{\textwidth}{p{1.7cm}X}
\toprule
Role & Content \\
\midrule

System &
You are a rigorous benchmark-label-leakage annotator for news--evidence pairs. Your task is to determine whether the candidate evidence directly reveals the benchmark authenticity label of the target news claim.

Judgment criteria:

Return \texttt{Leakage} if the evidence satisfies any of the following conditions:
(1) It directly or indirectly states a final verdict that the target news claim is true, false, fabricated, misleading, or a rumor.

(2) It reproduces a fact-checking conclusion, dataset annotation, benchmark label, or other annotation-specific information about the target claim.

(3) It contains an official denial, clarification, confirmation, judicial decision, or retrospective conclusion that itself functions as a complete verdict on the authenticity of the target claim, without requiring evidence comparison or reasoning.

Return \texttt{NoLeakage} if the evidence provides independent event facts,
background information, source statements, timelines, subsequent investigations, event developments, judicial outcomes, or official materials without directly stating the benchmark label or a complete final verdict.

Do not classify evidence as \texttt{Leakage} solely because it was published after the target news item. Post-publication factual evidence is admissible in the post-hoc fact-checking setting.

Judge only the supplied news claim and evidence. Do not use external knowledge
or infer missing facts.

Output requirement: Return exactly one label:
\texttt{Leakage} or \texttt{NoLeakage}. Do not provide explanations. \\

\midrule
User &
News claim: \texttt{[NEWS CLAIM]}

Candidate evidence: \texttt{[CANDIDATE EVIDENCE]} \\

\midrule
LLM &
\texttt{Leakage} or \texttt{NoLeakage} \\

\bottomrule
\end{tabularx}
\end{table*}

\subsection{Filtering}
We apply the same leakage-control procedure to both the pre-news and post-news evidence sets. In this work, benchmark-label leakage refers specifically to evidence that directly reveals the benchmark authenticity label, reproduces dataset annotations, or provides an explicit final verdict on the target claim without requiring evidence comparison or reasoning. The publication time of a piece of evidence alone is not used to determine whether it contains label leakage.

The first stage is automatic explicit filtering. We exclude target pages, same-domain reposts, near-duplicate copies, professional fact-checking pages, dataset annotation pages, and candidate webpages or snippets that directly state a verdict on the target claim. The explicit filtering rules cover expressions such as ``true,'' ``false,'' ``fake news,'' ``rumor,'' ``misinformation,'' ``misrepresentation,'' ``unverified,'' ``fact check,'' and ``debunked,'' together with their Chinese equivalents and semantically equivalent verdict-style expressions.

\begin{table*}[t]
\centering
\small
\caption{Temporal composition and leakage-control statistics of the retrieved evidence. The final automatic evidence set corresponds to the evidence retained after LLM screening.}
\label{tab:leakage_audit_stats}
\begin{tabular}{llrrr}
\toprule
Dataset & Temporal Subset & Candidates
& After Explicit Filtering
& After LLM Screening \\
\midrule
\multirow{2}{*}{Weibo-21}
& Pre-news  & 8,193  & 6,448  & 4,094 \\
& Post-news & 9,894  & 7,322  & 4,247 \\
\midrule
\multirow{2}{*}{Fakeddit}
& Pre-news  & 11,727 & 8,456  & 6,757 \\
& Post-news & 13,791 & 7,333  & 4,618 \\
\midrule
\multirow{2}{*}{SSS}
& Pre-news  & 18,526 & 12,324 & 8,606 \\
& Post-news & 21,718 & 9,167  & 6,551 \\
\bottomrule
\end{tabular}
\end{table*}

\begin{table*}[t]
\centering
\small
\caption{Full prompt used for LLM-based stance quantification.}
\label{tab:stance_prompt}
\begin{tabularx}{\textwidth}{X}
\toprule
System Prompt \\
\midrule
You are a rigorous semantic stance annotator for news-evidence pairs. You need to judge the semantic relation between the evidence and the core event, subject, and claim expressed by the news.

Please assign a continuous stance score in the interval $[0,1]$ to each evidence item. Interpret the score using the following ranges:

\begin{itemize}[leftmargin=1.6em,itemsep=1pt,topsep=2pt]
    \item $[0.0,0.2]$ Strong refutation: the evidence directly conflicts with the core claim of the news, or explicitly provides facts opposite to the core content of the news.
    \item $(0.2,0.4]$ Weak refutation: the evidence is partially inconsistent with the news, or weakly negates or weakens a key detail of the news.
    \item $(0.4,0.6]$ Neutral: the evidence is only background-related or topic-related, provides insufficient information, cannot determine support or refutation, or is basically irrelevant to the news.
    \item $(0.6,0.8]$ Weak support: the evidence is partially consistent with the news, or provides weak support for a key detail of the news.
    \item $(0.8,1.0]$ Strong support: the evidence explicitly supports the core claim of the news, or provides facts highly consistent with the core content of the news.
\end{itemize}

Important constraints:

\begin{itemize}[leftmargin=1.6em,itemsep=1pt,topsep=2pt]
    \item Support or refutation only indicates the semantic relation between the evidence text and the news text, and is independent of the actual truthfulness of the news.
    \item Do not complete missing facts using commonsense or external knowledge. Make the judgment only based on the input text.
\end{itemize}

Output requirement: Return only one decimal number in $[0,1]$, without explanations or additional text.
\\
\bottomrule
\end{tabularx}
\end{table*}

\begin{table*}[t]
\centering
\small
\caption{Mapping rules for the continuous stance scores output by the LLM.}
\label{tab:mapping_rules}
\begin{tabular}{c c p{0.65\textwidth}}
\toprule
 Score Range & Mapped Value & Meaning \\
\midrule
$[0.0, 0.2]$ & 0.1 & Severe conflict in core facts or direct denial. \\
\addlinespace[2pt]
$(0.2, 0.4]$ & 0.3 & Certain contradictions or inconsistencies exist, with weak refutation strength. \\
\addlinespace[2pt]
$(0.4, 0.6]$ & 0.5 & Content is irrelevant or lacks sufficient information to support a judgment. \\
\addlinespace[2pt]
$(0.6, 0.8]$ & 0.7 & Possesses partial factual consistency, providing limited support. \\
\addlinespace[2pt]
$(0.8, 1.0]$ & 0.9 & Highly consistent at the key factual level, explicitly supporting the original claim. \\
\bottomrule
\end{tabular}
\end{table*}

The second stage is LLM-based implicit leakage screening. This stage removes evidence that does not contain explicit verdict keywords but nevertheless directly discloses the benchmark label through a retrospective fact-checking conclusion, a dataset-like annotation, or an authoritative statement that itself constitutes a complete verdict on the target claim. Evidence identified as containing implicit label leakage is removed immediately.

By contrast, independent event facts are not regarded as label leakage merely because they were published after the news item. Subsequent investigations, official materials, source statements, and factual timelines may be retained when they provide evidence relevant to the claim without directly reproducing the benchmark label or an explicit final verdict. These post-publication facts constitute legitimate evidence in the post-hoc fact-checking setting represented by RAEGNet-A.

Table~\ref{tab:leakage_audit_stats} reports the temporal composition of the retrieved candidates and the number of evidence items retained after each automatic leakage-control stage. The pre-news and post-news subsets are filtered independently using the same criteria. The union of the two automatically retained subsets forms the evidence cache used by RAEGNet-A, whereas only the automatically retained pre-news subset is available to RAEGNet-B.

\subsection{Stance Quantification}
Only candidates judged to be leakage-free are used for stance quantification. According to the semantic direction between the evidence and the original news item, the LLM outputs an initial continuous stance score $\widetilde q_{ij}\in[0,1]$. Scores near the lower end of the interval indicate that the candidate evidence tends to refute the original news, scores near the upper end indicate support, and scores in the middle indicate an unclear stance, insufficient relevance, or inadequate information. We then map $\widetilde q_{ij}$ into the discrete stance set $q_{ij}\in\{0.1,0.3,0.5,0.7,0.9\}$ using equal-width intervals. The mapping rules and interpretations are provided in Table~\ref{tab:mapping_rules}.

\subsection{Post-hoc Validation of Evidence Filtering}
The evidence used for model training and evaluation is produced by the automatic ELERF pipeline, including explicit filtering, LLM-based implicit leakage screening, and stance quantification. After the evidence cache is constructed, we conduct a post-hoc validation to examine the quality of the automatic filtering process.

In this validation, annotators inspect retained news--evidence pairs and check whether the evidence is semantically relevant, whether it contains explicit or implicit benchmark-label leakage, and whether the assigned stance score is consistent with the evidence content. Annotators do not have access to the ground-truth authenticity label or the cross-validation fold assignment. Under the leakage definition used in this work, the validation does not identify retained evidence that directly discloses benchmark labels or final verdicts.

During validation, reviewers were instructed not to flag a candidate as leakage merely because it was published after the corresponding news item. For post-news candidates, the reviewers distinguished independent factual developments from direct benchmark-label disclosure. Subsequent investigations, official documents, judicial outcomes, and event developments were considered admissible when they supplied factual evidence that still required comparison with the target claim. Evidence was considered leakage when it directly reproduced the benchmark label, stated a complete final verdict on the target claim, or exposed dataset-specific annotation information.

This filtering protocol is introduced to keep the experimental evidence independent from benchmark answers. It should not be interpreted as implying that authoritative post-publication information is undesirable in real-world fact-checking. In operational post-hoc verification, official statements, investigation results, and judicial outcomes are often necessary and valuable evidence. Our experiments adopt a conservative protocol that removes direct verdict disclosure so that performance reflects evidence retrieval, comparison, and relational reasoning rather than access to benchmark answers.

\subsection{Distribution of Evidence Stances}

\begin{figure}[h]
    \centering
    \includegraphics[width=\columnwidth]{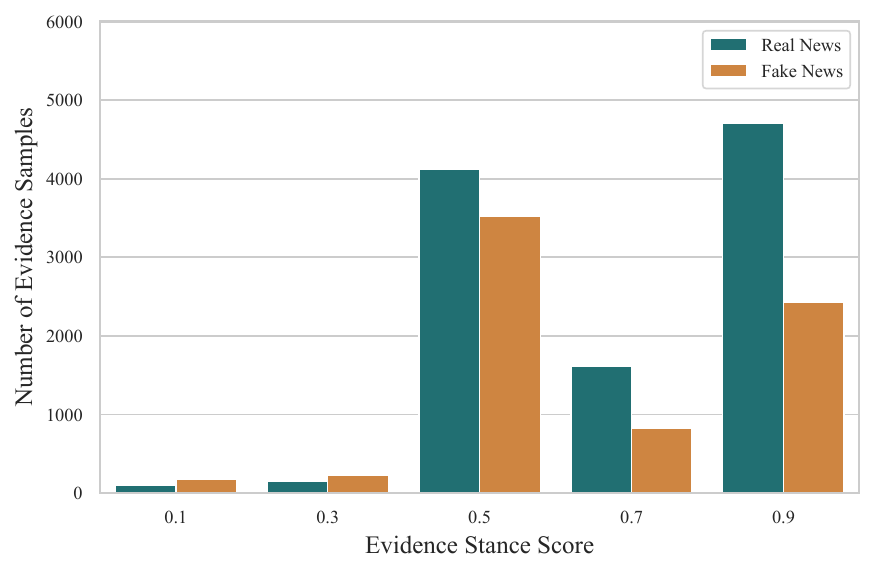}
    \caption{Distribution of external-evidence stance scores in the SSS dataset.}
    \label{fig:stance_distribution}
\end{figure}

To further validate the effectiveness of the evidence retrieval and stance quantification mechanisms, we analyze the stance distribution of external evidence associated with real and fake news in the SSS dataset. As shown in Figure~\ref{fig:stance_distribution}, real news is associated with substantially more retrieved evidence than fake news, and the number of evidence items corresponding to real news is markedly greater in the high-support interval. This result indicates that real-world events generally receive more extensive corroboration from external information.

\begin{figure*}[t]
    \centering
    \includegraphics[width=\textwidth]{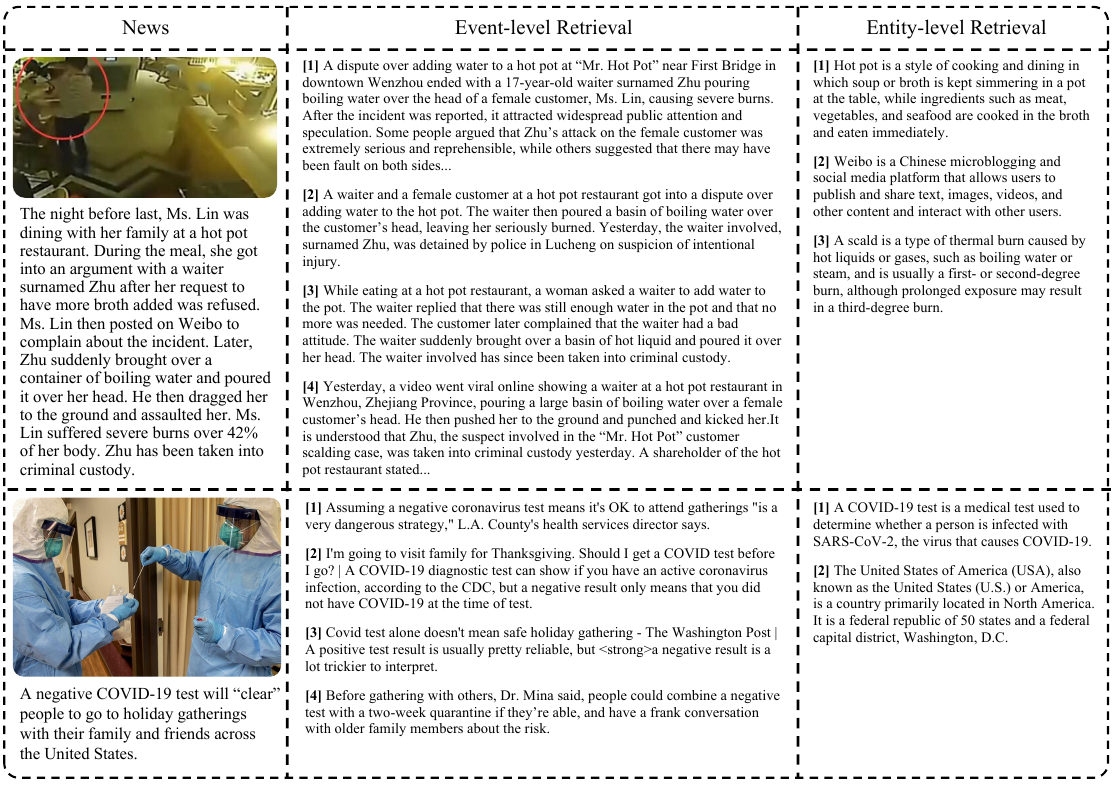}
    \caption{Qualitative cases comparing entity-level and event-level evidence retrieval. Event-level retrieval returns evidence that is more closely aligned with the complete news event, whereas entity-level retrieval tends to retrieve background information about salient entities mentioned in the news.}
    \label{fig:appendix_retrieval_cases}
\end{figure*}

\subsection{Evidence Settings and Intended Scenarios}
\label{sec:evidence_settings}

After automatic leakage-aware filtering, we construct two evidence settings corresponding to different information-availability scenarios.

\paragraph{RAEGNet-B: Early-detection Setting.}
RAEGNet-B uses only leakage-controlled evidence whose publication time precedes the corresponding news item. It therefore excludes all post-publication information and applies the complete benchmark-label-leakage-control procedure to the remaining pre-news evidence. This setting represents early fake-news detection under publication-date-based information-availability constraints.

\paragraph{RAEGNet-A: Post-hoc Fact-checking Setting.}
RAEGNet-A uses the union of the leakage-controlled pre-news and post-news evidence sets. It represents a post-hoc fact-checking scenario in which subsequent investigations, event developments, judicial outcomes, and official materials are legitimate sources of factual evidence. RAEGNet-A does not impose a temporal restriction, but it applies the same explicit filtering and LLM-based implicit screening to prevent direct disclosure of benchmark labels and final verdicts. It is therefore interpreted as evidence-supported verification after additional information becomes available.

The distinction between the two settings concerns information availability rather than filtering quality. RAEGNet-B evaluates whether event-level evidence is useful when only information available before news publication can be used, whereas RAEGNet-A evaluates evidence-supported verification when information emerging after publication is also available.

\subsection{Retrieval Case Analysis}
Figure~\ref{fig:appendix_retrieval_cases} provides qualitative cases illustrating how event-level retrieval differs from entity-level retrieval. These cases are not intended as a separate large-scale relevance audit, but they help reveal typical retrieval behaviors observed in our evidence collection process. In particular, entity-level queries often retrieve pages about salient persons, organizations, or locations, while event-level queries are more likely to retrieve documents describing the specific action, claim, or factual circumstance expressed by the news item.

\section{Statement on the Use of LLMs}
To ensure the transparency and reproducibility of this study, this section describes the large language models used in the experiments, their application scenarios, and their hyperparameter configurations.

\subsection{Model Selection for Core Modules}

\begin{itemize}[leftmargin=1.5em]
  \item Construction of the SSS dataset: We use the multimodal large language model Qwen2.5-VL-7B-Instruct. During data cleaning, the model jointly evaluates images and text to accurately remove low-quality samples.
  \item Evidence retrieval, implicit leakage screening, and stance quantification: We use Qwen2.5-VL-7B-Instruct to identify whether candidate evidence contains implicit label leakage and, only for leakage-free candidates, to assess the semantic relevance between candidate evidence and news items as well as the strength of support or refutation.
  \item LLM-based baselines: We use Qwen2.5-VL-7B-Instruct and InternVL2.5-8B. Qwen2.5-VL-7B-Instruct and InternVL2.5-8B receive the news text, image, and retrieved evidence snippets in a locally deployable zero-shot setting.
\end{itemize}

\subsection{Global Generation Parameters}

To ensure stable outputs, all model calls use a unified low-randomness configuration. The temperature is set to 0.3, nucleus sampling (top-$p$) is set to 0.85, and the maximum generation length is limited to 64 tokens. This configuration stabilizes the output format while also reducing API inference latency.

\section{Experimental Settings}
\label{sec:detailed_experimental_settings}

\subsection{Experimental Parameter Settings}
Table~\ref{tab:detailed_experimental_settings} lists the main hyperparameters used in RAEGNet.

\begin{table}[h]
\centering
\small
\caption{Detailed experimental settings.}
\label{tab:detailed_experimental_settings}
\begin{tabular}{lll}
\toprule
Type & Parameter & Value \\
\midrule

\multirow{5}{*}{Feature}
& Text dim. & 768 \\
& Image dim. & 768 \\
& Evidence dim. & 768 \\
& Projection dim. & 768 \\
& Relation dim. $d_r$ & 32 \\

\midrule
\multirow{8}{*}{Graph}
& $\delta_+$ & 0.6 \\
& $\delta_-$ & 0.4 \\
& $\delta_0$ & 0.1 \\
& $\kappa_{\mathrm{cor}}$ & 0.5 \\
& $\kappa_{\mathrm{con}}$ & 0.5 \\
& $\kappa_{\mathrm{sim}}$ & 0.5 \\
& GATv2 layers & 2 \\
& GATv2 heads & 1 \\

\midrule
\multirow{3}{*}{Loss}
& $\lambda_{\mathrm{harm}}$ & 0.25 \\
& $\lambda_{\mathrm{rank}}$ & 0.25 \\
& $\gamma$ & 1.00 \\

\midrule
\multirow{8}{*}{Training}
& Optimizer & AdamW \\
& Learning rate & $1\mathrm{e}{-4}$ \\
& Weight decay & 0.01 \\
& Batch size & 32 \\
& Max epochs & 50 \\
& Patience & 5 \\
& Dropout & 0.20 \\
& Random seed & 2026 \\

\bottomrule
\end{tabular}
\end{table}

\subsection{Baseline Models}
To evaluate the performance of RAEGNet, we compare it with several advanced methods. These baselines can be divided into three categories.

\paragraph{(1) Methods without external knowledge.} 
These methods focus on extracting features from the internal multimodal content of a news item and do not rely on external information.
\begin{itemize}[leftmargin=1.5em]
  \item CAFE \citep{chen2022cross}: CAFE evaluates cross-modal consistency through cross-modal ambiguity learning and adaptively aggregates unimodal and cross-modal features for detection.
  \item MRML \citep{peng2023mrml}: MRML uses triplet learning to discover inter-class relations within each modality and performs contrastive pairwise learning to model cross-modal relations.
  \item Event-Radar \citep{ma2024event}: Event-Radar proposes an event-driven multi-view learning framework that identifies event-irrelevant features to improve model generalization.
  \item MSACA \citep{wang2024fake}: MSACA uses a multi-scale semantic alignment mechanism to align and fuse text and image features at different semantic levels, thereby capturing fine-grained inconsistencies.
\end{itemize}

\paragraph{(2) Methods with external knowledge.}
These methods enhance reasoning by incorporating external knowledge graphs or background information.

\begin{itemize}[leftmargin=1.5em]
  \item KEHGNN-FD \citep{xie2023knowledge}: KEHGNN-FD uses a heterogeneous graph neural network to deeply integrate external structured knowledge with multimodal content features and capture rich semantic relations.
  \item NSLM \citep{dong2024unveiling}: NSLM proposes a neural-symbolic latent-variable model that combines news content with external evidence obtained through reverse-image retrieval. While determining news authenticity, it uncovers latent deceptive patterns to provide interpretability.
  \item ERIC-FND \citep{cao2025external}: ERIC-FND proposes an external-reliable-information-enhanced multimodal contrastive learning framework. It enriches news representations with entity-rich external information and uses multimodal contrastive learning to promote semantic interaction across modalities.
\end{itemize}

\begin{table*}[t]
\centering
\small
\caption{Full prompt used for direct w/ MLLM authenticity judgment.}
\label{tab:mllm_baseline_prompt}
\begin{tabularx}{\textwidth}{p{1.7cm}X}
\toprule
Role & Content \\
\midrule
System &
You are a rigorous multimodal news authenticity evaluator.
Your task is to determine whether a news item is fake or real by jointly considering the news text, the associated image, and the retrieved external evidence.

Decision criteria:

(1) Judge whether the core claim expressed by the news is supported or contradicted by the visual content and the retrieved evidence.

(2) Use the retrieved evidence only as contextual evidence. Do not assume that an evidence source is always correct, and do not use prior knowledge beyond the supplied inputs.

(3) If the image-text pair or the evidence reveals clear factual contradiction, temporal inconsistency, entity mismatch, or unsupported fabrication, classify the news as fake.

(4) If the news text, image, and evidence are mutually consistent and the evidence supports the core claim, classify the news as real.

(5) If the evidence is insufficient, make the most likely binary judgment based on the supplied news text, image, and evidence.

Output requirement: Return exactly one label:
\texttt{Fake} or \texttt{Real}. Do not provide explanations. \\

\midrule

User &
News text: \texttt{[NEWS TEXT]}

News image: \texttt{[NEWS IMAGE]}

Retrieved evidence snippets: \texttt{[EVIDENCE SNIPPETS]} \\

\midrule

MLLM &
\texttt{Fake} or \texttt{Real} \\

\bottomrule
\end{tabularx}
\end{table*}

\paragraph{(3) Methods with LLMs.}
These methods use multimodal or large language models to incorporate news content and external evidence for authenticity judgment.
\begin{itemize}[leftmargin=1.5em]
  \item Qwen2.5-VL-7B-Instruct: an open-source multimodal instruction model used as a general visual-language reasoning baseline.
  \item InternVL2.5-8B: an open-source multimodal model used to test direct evidence-conditioned visual-language judgment.
  \item GLPN-LLM \citep{hu2025synergizing}: GLPN-LLM combines large language models with global label propagation and uses LLM-generated pseudo-labels to alleviate the scarcity of supervised data.
\end{itemize}

\subsection{Prompt for w/ MLLM Baselines}
\label{sec:mllm_baseline_prompt}

For the direct w/ MLLM baselines, we use a fixed zero-shot prompt for the held-out samples in all five folds. The prompt asks the model to jointly consider the news text, news image, and retrieved evidence snippets, and to output only a binary authenticity label. GLPN-LLM follows its original LLM-based label-propagation setting.

\section{Complete Experimental Results}
\label{sec:complete_experimental_results}

This section provides the complete experimental results and additional analyses. To keep the presentation compact and readable, all textual discussions are placed before the tables, and the complete numerical results are collected at the end of this section.

We report three variants of RAEGNet to distinguish retrieval granularity and evidence availability. RAEGNet-E uses the entity-level evidence retrieved by ERIC-FND, serving as a controlled comparison with entity-level retrieval under the same model architecture. RAEGNet-B uses only leakage-controlled event-level evidence published before the corresponding news item and represents the early-detection setting under publication-date-based evidence availability. RAEGNet-A uses the union of leakage-controlled pre-news and post-news event-level evidence and represents the post-hoc fact-checking setting. Post-publication independent factual evidence is admissible in RAEGNet-A, whereas evidence that directly reveals the benchmark label, reproduces dataset annotations, or states a complete final verdict is excluded from both settings.

\subsection{Overall Performance}
\label{sec:complete_overall_performance}

Tables~\ref{tab:complete_results_weibo}--\ref{tab:complete_results_sss} report the overall performance on the three datasets. Across datasets, RAEGNet consistently achieves stronger results than task-specific baselines, external-knowledge-enhanced baselines, and direct LLM-based baselines. The comparison among RAEGNet-E, RAEGNet-B, and RAEGNet-A further shows that event-level evidence is more effective than entity-level evidence under the same architecture, while broader leakage-controlled evidence provides additional gains in the post-hoc setting.

\subsection{Ablation Study}
\label{sec:complete_ablation}

Tables~\ref{tab:ablation_weibo}--\ref{tab:ablation_sss} present the complete ablation results. Removing the evidence graph, edge weights, or edge types weakens the model, confirming the importance of relation-aware evidence modeling. Removing the harm-aware branch or its associated objectives mainly affects high-harm fake-news recognition, showing that the harm-aware design contributes beyond ordinary authenticity classification. Removing text, image, or evidence also reduces performance, indicating that the final model benefits from all three information sources.

\subsection{Performance Across Harm Levels}
\label{sec:complete_harm_levels}

Tables~\ref{tab:harm_weibo}--\ref{tab:harm_sss} report accuracy across different harm levels. The results show that RAEGNet maintains stable performance across harm groups and is particularly effective on high-harm samples. This supports the central motivation of the model: fake-news detection should not only optimize aggregate performance, but should also strengthen recognition of more socially consequential misinformation.

\subsection{Conditional-Harm Prediction}
\label{sec:conditional_harm_prediction}

Table~\ref{tab:harm_prediction} directly evaluates the conditional-harm prediction branch. Compared with a text-only predictor, the RAEGNet harm branch better matches human harm annotations, indicating that the multimodal representation and the learned harm-risk branch capture useful signals for estimating potential social impact.

\subsection{Harm-Matched Evaluation}
\label{sec:harm_matched_evaluation}

Table~\ref{tab:harm_matched_evaluation} reports the harm-matched evaluation, where real and fake samples are balanced within each harm level. This setting reduces the possibility that the model benefits merely from marginal correlations between harm labels and authenticity labels. The full model remains consistently better than the ablated variant, suggesting that the harm-aware objective improves detection through useful supervision rather than simply exploiting label distribution patterns.

\subsection{Analysis of Retrieval and Model Contributions}
\label{sec:retrieval_model_contribution}

Table~\ref{tab:controlled_retrieval_model} separates the effects of retrieval granularity and downstream model architecture under the same source pool, filtering process, and evidence budget. Event-level retrieval improves over entity-level retrieval under the same backbone, showing that complete-event queries reduce irrelevant evidence. Under the same retrieval setting, RAEGNet improves over simple evidence fusion, demonstrating that relation-aware graph modeling contributes additional gains beyond retrieval quality alone.

\subsection{Event-Level Evidence for Other w/ EK Models}
\label{sec:ek_event_evidence_transfer}

Table~\ref{tab:ek_event_evidence_transfer} evaluates whether ELERF evidence benefits other external-knowledge baselines. Replacing the original evidence sources with leakage-filtered ELERF evidence consistently improves these baselines, especially entity-oriented methods. This indicates that the advantage of ELERF comes not only from its compatibility with RAEGNet, but also from the general usefulness of event-level evidence for evidence-enhanced fake-news detection.

\clearpage

\begin{table*}[t]
\centering
\footnotesize
\setlength{\tabcolsep}{4.2pt}
\renewcommand{\arraystretch}{1.08}
\caption{Overall performance on Weibo-21.}
\label{tab:complete_results_weibo}
\begin{tabular}{clccccccc}
\toprule
\multirow{2}{*}{Category}
& \multirow{2}{*}{Method}
& \multirow{2}{*}{Accuracy}
& \multicolumn{3}{c}{Fake News}
& \multicolumn{3}{c}{Real News} \\
\cmidrule(lr){4-6}
\cmidrule(lr){7-9}
& & & Precision & Recall & F1
& Precision & Recall & F1 \\
\midrule
\multirow{4}{*}{w/o EK}
& CAFE & \meanstd{0.815}{0.005} & \meanstd{0.850}{0.005} & \meanstd{0.774}{0.005} & \meanstd{0.810}{0.005} & \meanstd{0.785}{0.005} & \meanstd{0.858}{0.005} & \meanstd{0.820}{0.005} \\
& MRML & \meanstd{0.903}{0.011} & \meanstd{0.881}{0.010} & \meanstd{0.935}{0.010} & \meanstd{0.908}{0.010} & \meanstd{0.928}{0.012} & \meanstd{0.869}{0.011} & \meanstd{0.898}{0.011} \\
& Event-Radar & \meanstd{0.881}{0.010} & \meanstd{0.873}{0.009} & \meanstd{0.896}{0.010} & \meanstd{0.884}{0.009} & \meanstd{0.889}{0.010} & \meanstd{0.865}{0.010} & \meanstd{0.877}{0.010} \\
& MSACA & \meanstd{0.894}{0.006} & \meanstd{0.861}{0.006} & \meanstd{0.944}{0.005} & \meanstd{0.900}{0.005} & \meanstd{0.935}{0.005} & \meanstd{0.841}{0.007} & \meanstd{0.886}{0.006} \\
\midrule
\multirow{3}{*}{w/ EK}
& KEHGNN-FD & \meanstd{0.764}{0.005} & \meanstd{0.752}{0.004} & \meanstd{0.801}{0.005} & \meanstd{0.776}{0.004} & \meanstd{0.779}{0.005} & \meanstd{0.726}{0.005} & \meanstd{0.751}{0.005} \\
& NSLM & \meanstd{0.878}{0.009} & \meanstd{0.851}{0.008} & \meanstd{0.921}{0.009} & \meanstd{0.885}{0.008} & \meanstd{0.911}{0.010} & \meanstd{0.833}{0.009} & \meanstd{0.870}{0.010} \\
& ERIC-FND & \meanstd{0.844}{0.011} & \meanstd{0.804}{0.009} & \meanstd{0.916}{0.011} & \meanstd{0.857}{0.010} & \meanstd{0.898}{0.013} & \meanstd{0.769}{0.011} & \meanstd{0.828}{0.012} \\
\midrule
\multirow{3}{*}{w/ LLM}
& Qwen2.5-VL & \meanstd{0.742}{0.010} & \meanstd{0.734}{0.009} & \meanstd{0.776}{0.008} & \meanstd{0.754}{0.009} & \meanstd{0.753}{0.010} & \meanstd{0.707}{0.011} & \meanstd{0.729}{0.010} \\
& InternVL2.5 & \meanstd{0.714}{0.007} & \meanstd{0.709}{0.007} & \meanstd{0.744}{0.007} & \meanstd{0.726}{0.007} & \meanstd{0.720}{0.008} & \meanstd{0.683}{0.007} & \meanstd{0.701}{0.007} \\
& GLPN-LLM & \meanstd{0.882}{0.010} & \meanstd{0.856}{0.010} & \meanstd{0.923}{0.010} & \meanstd{0.888}{0.010} & \meanstd{0.913}{0.011} & \meanstd{0.839}{0.011} & \meanstd{0.874}{0.011} \\
\midrule
\multirow{3}{*}{Ours}
& RAEGNet-E & \meanstd{0.911}{0.004} & \meanstd{0.928}{0.004} & \meanstd{0.896}{0.004} & \meanstd{0.911}{0.004} & \meanstd{0.895}{0.004} & \meanstd{0.928}{0.004} & \meanstd{0.911}{0.004} \\
& RAEGNet-B & \meanstd{0.928}{0.007} & \meanstd{0.923}{0.007} & \meanstd{0.937}{0.007} & \meanstd{0.930}{0.007} & \meanstd{0.933}{0.007} & \meanstd{0.918}{0.007} & \meanstd{0.926}{0.007} \\
& RAEGNet-A & \meanstd{0.931}{0.009} & \meanstd{0.919}{0.008} & \meanstd{0.948}{0.009} & \meanstd{0.934}{0.009} & \meanstd{0.945}{0.010} & \meanstd{0.913}{0.009} & \meanstd{0.929}{0.009} \\
\bottomrule
\end{tabular}
\end{table*}

\begin{table*}[t]
\centering
\footnotesize
\setlength{\tabcolsep}{4.2pt}
\renewcommand{\arraystretch}{1.08}
\caption{Overall performance on Fakeddit.}
\label{tab:complete_results_fakeddit}
\begin{tabular}{clccccccc}
\toprule
\multirow{2}{*}{Category}
& \multirow{2}{*}{Method}
& \multirow{2}{*}{Accuracy}
& \multicolumn{3}{c}{Fake News}
& \multicolumn{3}{c}{Real News} \\
\cmidrule(lr){4-6}
\cmidrule(lr){7-9}
& & & Precision & Recall & F1
& Precision & Recall & F1 \\
\midrule
\multirow{4}{*}{w/o EK}
& CAFE & \meanstd{0.786}{0.011} & \meanstd{0.773}{0.009} & \meanstd{0.864}{0.011} & \meanstd{0.816}{0.010} & \meanstd{0.807}{0.015} & \meanstd{0.691}{0.012} & \meanstd{0.745}{0.013} \\
& MRML & \meanstd{0.860}{0.006} & \meanstd{0.852}{0.005} & \meanstd{0.900}{0.006} & \meanstd{0.876}{0.005} & \meanstd{0.870}{0.007} & \meanstd{0.811}{0.007} & \meanstd{0.839}{0.007} \\
& Event-Radar & \meanstd{0.840}{0.010} & \meanstd{0.831}{0.008} & \meanstd{0.888}{0.010} & \meanstd{0.859}{0.009} & \meanstd{0.852}{0.012} & \meanstd{0.780}{0.010} & \meanstd{0.815}{0.011} \\
& MSACA & \meanstd{0.846}{0.009} & \meanstd{0.839}{0.008} & \meanstd{0.889}{0.008} & \meanstd{0.864}{0.008} & \meanstd{0.855}{0.010} & \meanstd{0.794}{0.010} & \meanstd{0.823}{0.010} \\
\midrule
\multirow{3}{*}{w/ EK}
& KEHGNN-FD & \meanstd{0.763}{0.010} & \meanstd{0.752}{0.008} & \meanstd{0.846}{0.009} & \meanstd{0.796}{0.008} & \meanstd{0.780}{0.013} & \meanstd{0.662}{0.011} & \meanstd{0.716}{0.012} \\
& NSLM & \meanstd{0.850}{0.006} & \meanstd{0.844}{0.005} & \meanstd{0.892}{0.006} & \meanstd{0.867}{0.005} & \meanstd{0.859}{0.008} & \meanstd{0.800}{0.006} & \meanstd{0.828}{0.007} \\
& ERIC-FND & \meanstd{0.794}{0.010} & \meanstd{0.764}{0.008} & \meanstd{0.904}{0.010} & \meanstd{0.828}{0.009} & \meanstd{0.851}{0.015} & \meanstd{0.661}{0.011} & \meanstd{0.744}{0.013} \\
\midrule
\multirow{3}{*}{w/ LLM}
& Qwen2.5-VL & \meanstd{0.738}{0.001} & \meanstd{0.740}{0.002} & \meanstd{0.805}{0.001} & \meanstd{0.771}{0.000} & \meanstd{0.736}{0.001} & \meanstd{0.657}{0.003} & \meanstd{0.694}{0.002} \\
& InternVL2.5 & \meanstd{0.726}{0.000} & \meanstd{0.726}{0.000} & \meanstd{0.804}{0.000} & \meanstd{0.763}{0.000} & \meanstd{0.727}{0.000} & \meanstd{0.632}{0.000} & \meanstd{0.676}{0.000} \\
& GLPN-LLM & \meanstd{0.875}{0.005} & \meanstd{0.871}{0.004} & \meanstd{0.906}{0.004} & \meanstd{0.888}{0.004} & \meanstd{0.880}{0.006} & \meanstd{0.837}{0.005} & \meanstd{0.858}{0.005} \\
\midrule
\multirow{3}{*}{Ours}
& RAEGNet-E & \meanstd{0.898}{0.007} & \meanstd{0.898}{0.007} & \meanstd{0.919}{0.006} & \meanstd{0.908}{0.007} & \meanstd{0.898}{0.008} & \meanstd{0.873}{0.009} & \meanstd{0.886}{0.008} \\
& RAEGNet-B & \meanstd{0.911}{0.004} & \meanstd{0.893}{0.004} & \meanstd{0.953}{0.004} & \meanstd{0.922}{0.004} & \meanstd{0.937}{0.005} & \meanstd{0.861}{0.005} & \meanstd{0.898}{0.005} \\
& RAEGNet-A & \meanstd{0.919}{0.007} & \meanstd{0.921}{0.006} & \meanstd{0.932}{0.006} & \meanstd{0.926}{0.006} & \meanstd{0.916}{0.008} & \meanstd{0.903}{0.007} & \meanstd{0.910}{0.007} \\
\bottomrule
\end{tabular}
\end{table*}

\begin{table*}[t]
\centering
\footnotesize
\setlength{\tabcolsep}{4.2pt}
\renewcommand{\arraystretch}{1.08}
\caption{Overall performance on SSS.}
\label{tab:complete_results_sss}
\begin{tabular}{clccccccc}
\toprule
\multirow{2}{*}{Category}
& \multirow{2}{*}{Method}
& \multirow{2}{*}{Accuracy}
& \multicolumn{3}{c}{Fake News}
& \multicolumn{3}{c}{Real News} \\
\cmidrule(lr){4-6}
\cmidrule(lr){7-9}
& & & Precision & Recall & F1
& Precision & Recall & F1 \\
\midrule
\multirow{4}{*}{w/o EK}
& CAFE & \meanstd{0.726}{0.009} & \meanstd{0.709}{0.008} & \meanstd{0.802}{0.009} & \meanstd{0.753}{0.009} & \meanstd{0.750}{0.012} & \meanstd{0.644}{0.010} & \meanstd{0.693}{0.010} \\
& MRML & \meanstd{0.758}{0.006} & \meanstd{0.742}{0.005} & \meanstd{0.821}{0.006} & \meanstd{0.779}{0.005} & \meanstd{0.780}{0.007} & \meanstd{0.690}{0.006} & \meanstd{0.732}{0.006} \\
& Event-Radar & \meanstd{0.754}{0.005} & \meanstd{0.755}{0.005} & \meanstd{0.779}{0.005} & \meanstd{0.767}{0.005} & \meanstd{0.752}{0.006} & \meanstd{0.726}{0.005} & \meanstd{0.739}{0.006} \\
& MSACA & \meanstd{0.762}{0.011} & \meanstd{0.757}{0.010} & \meanstd{0.801}{0.010} & \meanstd{0.778}{0.010} & \meanstd{0.769}{0.012} & \meanstd{0.721}{0.012} & \meanstd{0.744}{0.012} \\
\midrule
\multirow{3}{*}{w/ EK}
& KEHGNN-FD & \meanstd{0.674}{0.010} & \meanstd{0.672}{0.009} & \meanstd{0.726}{0.010} & \meanstd{0.698}{0.009} & \meanstd{0.675}{0.011} & \meanstd{0.616}{0.010} & \meanstd{0.644}{0.011} \\
& NSLM & \meanstd{0.781}{0.010} & \meanstd{0.751}{0.009} & \meanstd{0.866}{0.010} & \meanstd{0.804}{0.009} & \meanstd{0.825}{0.013} & \meanstd{0.688}{0.011} & \meanstd{0.750}{0.012} \\
& ERIC-FND & \meanstd{0.655}{0.009} & \meanstd{0.620}{0.006} & \meanstd{0.870}{0.007} & \meanstd{0.724}{0.007} & \meanstd{0.749}{0.015} & \meanstd{0.421}{0.010} & \meanstd{0.539}{0.013} \\
\midrule
\multirow{3}{*}{w/ LLM}
& Qwen2.5-VL & \meanstd{0.629}{0.005} & \meanstd{0.628}{0.005} & \meanstd{0.703}{0.005} & \meanstd{0.664}{0.005} & \meanstd{0.630}{0.006} & \meanstd{0.549}{0.006} & \meanstd{0.587}{0.006} \\
& InternVL2.5 & \meanstd{0.613}{0.005} & \meanstd{0.614}{0.005} & \meanstd{0.690}{0.004} & \meanstd{0.650}{0.004} & \meanstd{0.612}{0.006} & \meanstd{0.530}{0.007} & \meanstd{0.568}{0.007} \\
& GLPN-LLM & \meanstd{0.707}{0.007} & \meanstd{0.684}{0.006} & \meanstd{0.814}{0.006} & \meanstd{0.743}{0.006} & \meanstd{0.746}{0.009} & \meanstd{0.592}{0.007} & \meanstd{0.660}{0.008} \\
\midrule
\multirow{3}{*}{Ours}
& RAEGNet-E & \meanstd{0.808}{0.010} & \meanstd{0.810}{0.009} & \meanstd{0.824}{0.010} & \meanstd{0.817}{0.010} & \meanstd{0.805}{0.011} & \meanstd{0.790}{0.010} & \meanstd{0.798}{0.010} \\
& RAEGNet-B & \meanstd{0.837}{0.011} & \meanstd{0.810}{0.010} & \meanstd{0.897}{0.011} & \meanstd{0.851}{0.010} & \meanstd{0.873}{0.013} & \meanstd{0.772}{0.012} & \meanstd{0.819}{0.013} \\
& RAEGNet-A & \meanstd{0.843}{0.012} & \meanstd{0.803}{0.010} & \meanstd{0.926}{0.011} & \meanstd{0.860}{0.011} & \meanstd{0.903}{0.015} & \meanstd{0.754}{0.012} & \meanstd{0.822}{0.013} \\
\bottomrule
\end{tabular}
\end{table*}

\begin{table*}[t]
\centering
\small
\setlength{\tabcolsep}{7pt}
\renewcommand{\arraystretch}{1.10}
\caption{Ablation results on Weibo-21.}
\label{tab:ablation_weibo}
\begin{tabular}{lccccc}
\toprule
Variation & Accuracy & Macro F1 & Fake F1 & HHF F1 & HHF Recall \\
\midrule
w/o Evidence Graph & \meanstd{0.913}{0.007} & \meanstd{0.913}{0.007} & \meanstd{0.915}{0.007} & \meanstd{0.956}{0.007} & \meanstd{0.961}{0.007} \\
\quad -- w/o Edge Weights & \meanstd{0.920}{0.009} & \meanstd{0.920}{0.009} & \meanstd{0.921}{0.009} & \meanstd{0.959}{0.009} & \meanstd{0.956}{0.009} \\
\quad -- w/o Edge Types & \meanstd{0.914}{0.006} & \meanstd{0.914}{0.006} & \meanstd{0.916}{0.006} & \meanstd{0.960}{0.006} & \meanstd{0.967}{0.006} \\
\midrule
w/o Harm-aware Branch & \meanstd{0.915}{0.006} & \meanstd{0.915}{0.006} & \meanstd{0.914}{0.005} & \meanstd{0.946}{0.007} & \meanstd{0.919}{0.006} \\
\quad -- w/o $\mathcal{L}_{\mathrm{harm}}$ & \meanstd{0.923}{0.010} & \meanstd{0.923}{0.010} & \meanstd{0.923}{0.011} & \meanstd{0.954}{0.010} & \meanstd{0.936}{0.011} \\
\quad -- w/o $\mathcal{L}_{\mathrm{rank}}$ & \meanstd{0.924}{0.007} & \meanstd{0.924}{0.007} & \meanstd{0.924}{0.008} & \meanstd{0.947}{0.008} & \meanstd{0.932}{0.009} \\
\midrule
w/o News Text & \meanstd{0.886}{0.004} & \meanstd{0.886}{0.004} & \meanstd{0.888}{0.004} & \meanstd{0.919}{0.005} & \meanstd{0.891}{0.004} \\
w/o News Image & \meanstd{0.859}{0.012} & \meanstd{0.859}{0.012} & \meanstd{0.865}{0.012} & \meanstd{0.925}{0.011} & \meanstd{0.919}{0.013} \\
w/o Evidence & \meanstd{0.907}{0.006} & \meanstd{0.907}{0.006} & \meanstd{0.909}{0.006} & \meanstd{0.954}{0.007} & \meanstd{0.933}{0.007} \\
\midrule
Full Model & \meanstd{0.931}{0.009} & \meanstd{0.931}{0.009} & \meanstd{0.934}{0.009} & \meanstd{0.972}{0.006} & \meanstd{0.974}{0.007} \\
\bottomrule
\end{tabular}
\end{table*}

\begin{table*}[t]
\centering
\small
\setlength{\tabcolsep}{7pt}
\renewcommand{\arraystretch}{1.10}
\caption{Ablation results on Fakeddit.}
\label{tab:ablation_fakeddit}
\begin{tabular}{lccccc}
\toprule
Variation & Accuracy & Macro F1 & Fake F1 & HHF F1 & HHF Recall \\
\midrule
w/o Evidence Graph & \meanstd{0.909}{0.012} & \meanstd{0.907}{0.012} & \meanstd{0.921}{0.012} & \meanstd{0.939}{0.012} & \meanstd{0.952}{0.014} \\
\quad -- w/o Edge Weights & \meanstd{0.915}{0.004} & \meanstd{0.914}{0.004} & \meanstd{0.924}{0.004} & \meanstd{0.936}{0.005} & \meanstd{0.935}{0.005} \\
\quad -- w/o Edge Types & \meanstd{0.911}{0.011} & \meanstd{0.910}{0.011} & \meanstd{0.921}{0.011} & \meanstd{0.930}{0.011} & \meanstd{0.935}{0.011} \\
\midrule
w/o Harm-aware Branch & \meanstd{0.912}{0.010} & \meanstd{0.911}{0.010} & \meanstd{0.920}{0.010} & \meanstd{0.925}{0.011} & \meanstd{0.907}{0.009} \\
\quad -- w/o $\mathcal{L}_{\mathrm{harm}}$ & \meanstd{0.909}{0.010} & \meanstd{0.908}{0.010} & \meanstd{0.917}{0.010} & \meanstd{0.926}{0.010} & \meanstd{0.919}{0.009} \\
\quad -- w/o $\mathcal{L}_{\mathrm{rank}}$ & \meanstd{0.906}{0.007} & \meanstd{0.905}{0.007} & \meanstd{0.913}{0.007} & \meanstd{0.922}{0.007} & \meanstd{0.900}{0.007} \\
\midrule
w/o News Text & \meanstd{0.870}{0.011} & \meanstd{0.867}{0.011} & \meanstd{0.888}{0.011} & \meanstd{0.894}{0.011} & \meanstd{0.935}{0.011} \\
w/o News Image & \meanstd{0.860}{0.005} & \meanstd{0.857}{0.005} & \meanstd{0.878}{0.005} & \meanstd{0.894}{0.005} & \meanstd{0.943}{0.005} \\
w/o Evidence & \meanstd{0.886}{0.011} & \meanstd{0.885}{0.011} & \meanstd{0.898}{0.011} & \meanstd{0.929}{0.012} & \meanstd{0.942}{0.009} \\
\midrule
Full Model & \meanstd{0.919}{0.007} & \meanstd{0.918}{0.007} & \meanstd{0.926}{0.006} & \meanstd{0.947}{0.006} & \meanstd{0.955}{0.005} \\
\bottomrule
\end{tabular}
\end{table*}

\begin{table*}[t]
\centering
\small
\setlength{\tabcolsep}{7pt}
\renewcommand{\arraystretch}{1.10}
\caption{Ablation results on SSS.}
\label{tab:ablation_sss}
\begin{tabular}{lccccc}
\toprule
Variation & Accuracy & Macro F1 & Fake F1 & HHF F1 & HHF Recall \\
\midrule
w/o Evidence Graph & \meanstd{0.822}{0.004} & \meanstd{0.818}{0.004} & \meanstd{0.844}{0.005} & \meanstd{0.892}{0.004} & \meanstd{0.956}{0.005} \\
\quad -- w/o Edge Weights & \meanstd{0.837}{0.011} & \meanstd{0.835}{0.010} & \meanstd{0.851}{0.011} & \meanstd{0.886}{0.010} & \meanstd{0.923}{0.012} \\
\quad -- w/o Edge Types & \meanstd{0.826}{0.011} & \meanstd{0.824}{0.011} & \meanstd{0.843}{0.010} & \meanstd{0.877}{0.011} & \meanstd{0.928}{0.011} \\
\midrule
w/o Harm-aware Branch & \meanstd{0.839}{0.010} & \meanstd{0.838}{0.010} & \meanstd{0.850}{0.010} & \meanstd{0.878}{0.010} & \meanstd{0.905}{0.010} \\
\quad -- w/o $\mathcal{L}_{\mathrm{harm}}$ & \meanstd{0.839}{0.006} & \meanstd{0.838}{0.006} & \meanstd{0.852}{0.007} & \meanstd{0.888}{0.007} & \meanstd{0.923}{0.007} \\
\quad -- w/o $\mathcal{L}_{\mathrm{rank}}$ & \meanstd{0.833}{0.004} & \meanstd{0.832}{0.004} & \meanstd{0.844}{0.004} & \meanstd{0.892}{0.005} & \meanstd{0.917}{0.004} \\
\midrule
w/o News Text & \meanstd{0.815}{0.005} & \meanstd{0.814}{0.005} & \meanstd{0.830}{0.005} & \meanstd{0.891}{0.005} & \meanstd{0.910}{0.005} \\
w/o News Image & \meanstd{0.803}{0.006} & \meanstd{0.801}{0.006} & \meanstd{0.822}{0.006} & \meanstd{0.883}{0.006} & \meanstd{0.927}{0.007} \\
w/o Evidence & \meanstd{0.816}{0.008} & \meanstd{0.815}{0.008} & \meanstd{0.831}{0.008} & \meanstd{0.888}{0.008} & \meanstd{0.904}{0.009} \\
\midrule
Full Model & \meanstd{0.843}{0.012} & \meanstd{0.841}{0.012} & \meanstd{0.860}{0.011} & \meanstd{0.896}{0.011} & \meanstd{0.952}{0.011} \\
\bottomrule
\end{tabular}
\end{table*}

\begin{table*}[t]
\centering
\small
\setlength{\tabcolsep}{5.5pt}
\renewcommand{\arraystretch}{1.08}
\caption{Accuracy across harm levels on Weibo-21.}
\label{tab:harm_weibo}
\begin{tabular}{clcccccc}
\toprule
Category & Method & 0.1 & 0.3 & 0.5 & 0.7 & 0.9 & Overall \\
\midrule
\multirow{4}{*}{w/o EK}
& CAFE & \meanstd{0.825}{0.011} & \meanstd{0.806}{0.004} & \meanstd{0.816}{0.008} & \meanstd{0.805}{0.006} & \meanstd{0.818}{0.008} & \meanstd{0.815}{0.005} \\
& MRML & \meanstd{0.898}{0.011} & \meanstd{0.909}{0.009} & \meanstd{0.902}{0.012} & \meanstd{0.908}{0.011} & \meanstd{0.900}{0.008} & \meanstd{0.903}{0.011} \\
& Event-Radar & \meanstd{0.886}{0.012} & \meanstd{0.871}{0.005} & \meanstd{0.882}{0.010} & \meanstd{0.879}{0.012} & \meanstd{0.884}{0.008} & \meanstd{0.881}{0.010} \\
& MSACA & \meanstd{0.889}{0.009} & \meanstd{0.898}{0.007} & \meanstd{0.892}{0.005} & \meanstd{0.900}{0.006} & \meanstd{0.890}{0.001} & \meanstd{0.894}{0.006} \\
\midrule
\multirow{3}{*}{w/ EK}
& KEHGNN-FD & \meanstd{0.771}{0.009} & \meanstd{0.756}{0.005} & \meanstd{0.769}{0.008} & \meanstd{0.755}{0.007} & \meanstd{0.765}{0.002} & \meanstd{0.764}{0.005} \\
& NSLM & \meanstd{0.882}{0.009} & \meanstd{0.871}{0.010} & \meanstd{0.879}{0.010} & \meanstd{0.876}{0.009} & \meanstd{0.881}{0.009} & \meanstd{0.878}{0.009} \\
& ERIC-FND & \meanstd{0.838}{0.012} & \meanstd{0.850}{0.012} & \meanstd{0.843}{0.011} & \meanstd{0.847}{0.009} & \meanstd{0.843}{0.011} & \meanstd{0.844}{0.011} \\
\midrule
\multirow{3}{*}{w/ LLM}
& Qwen2.5-VL & \meanstd{0.746}{0.011} & \meanstd{0.739}{0.009} & \meanstd{0.744}{0.008} & \meanstd{0.739}{0.010} & \meanstd{0.743}{0.008} & \meanstd{0.742}{0.009} \\
& InternVL2.5 & \meanstd{0.722}{0.002} & \meanstd{0.715}{0.001} & \meanstd{0.719}{0.003} & \meanstd{0.715}{0.003} & \meanstd{0.724}{0.008} & \meanstd{0.714}{0.007} \\
& GLPN-LLM & \meanstd{0.884}{0.010} & \meanstd{0.878}{0.012} & \meanstd{0.884}{0.010} & \meanstd{0.879}{0.012} & \meanstd{0.884}{0.008} & \meanstd{0.882}{0.010} \\
\midrule
\multirow{3}{*}{Ours}
& RAEGNet-E & \meanstd{0.910}{0.010} & \meanstd{0.883}{0.005} & \meanstd{0.914}{0.003} & \meanstd{0.919}{0.005} & \meanstd{0.969}{0.000} & \meanstd{0.911}{0.004} \\
& RAEGNet-B & \meanstd{0.928}{0.007} & \meanstd{0.901}{0.006} & \meanstd{0.928}{0.009} & \meanstd{0.931}{0.007} & \meanstd{0.991}{0.009} & \meanstd{0.928}{0.007} \\
& RAEGNet-A & \meanstd{0.931}{0.011} & \meanstd{0.905}{0.012} & \meanstd{0.928}{0.009} & \meanstd{0.938}{0.009} & \meanstd{0.997}{0.007} & \meanstd{0.931}{0.009} \\
\bottomrule
\end{tabular}
\end{table*}

\begin{table*}[t]
\centering
\small
\setlength{\tabcolsep}{5.5pt}
\renewcommand{\arraystretch}{1.08}
\caption{Accuracy across harm levels on Fakeddit.}
\label{tab:harm_fakeddit}
\begin{tabular}{clcccccc}
\toprule
Category & Method & 0.1 & 0.3 & 0.5 & 0.7 & 0.9 & Overall \\
\midrule
\multirow{4}{*}{w/o EK}
& CAFE & \meanstd{0.792}{0.011} & \meanstd{0.782}{0.012} & \meanstd{0.789}{0.011} & \meanstd{0.783}{0.012} & \meanstd{0.786}{0.010} & \meanstd{0.786}{0.011} \\
& MRML & \meanstd{0.852}{0.008} & \meanstd{0.865}{0.009} & \meanstd{0.858}{0.009} & \meanstd{0.863}{0.002} & \meanstd{0.860}{0.001} & \meanstd{0.860}{0.006} \\
& Event-Radar & \meanstd{0.844}{0.008} & \meanstd{0.836}{0.011} & \meanstd{0.842}{0.011} & \meanstd{0.837}{0.011} & \meanstd{0.839}{0.006} & \meanstd{0.840}{0.010} \\
& MSACA & \meanstd{0.840}{0.010} & \meanstd{0.849}{0.008} & \meanstd{0.844}{0.011} & \meanstd{0.850}{0.011} & \meanstd{0.847}{0.001} & \meanstd{0.846}{0.009} \\
\midrule
\multirow{3}{*}{w/ EK}
& KEHGNN-FD & \meanstd{0.765}{0.007} & \meanstd{0.758}{0.011} & \meanstd{0.766}{0.012} & \meanstd{0.761}{0.012} & \meanstd{0.763}{0.006} & \meanstd{0.763}{0.010} \\
& NSLM & \meanstd{0.844}{0.011} & \meanstd{0.855}{0.010} & \meanstd{0.847}{0.006} & \meanstd{0.854}{0.011} & \meanstd{0.849}{0.005} & \meanstd{0.850}{0.006} \\
& ERIC-FND & \meanstd{0.801}{0.010} & \meanstd{0.789}{0.011} & \meanstd{0.798}{0.010} & \meanstd{0.790}{0.011} & \meanstd{0.796}{0.009} & \meanstd{0.794}{0.010} \\
\midrule
\multirow{3}{*}{w/ LLM}
& Qwen2.5-VL & \meanstd{0.733}{0.003} & \meanstd{0.743}{0.001} & \meanstd{0.737}{0.002} & \meanstd{0.741}{0.001} & \meanstd{0.732}{0.002} & \meanstd{0.738}{0.001} \\
& InternVL2.5 & \meanstd{0.721}{0.000} & \meanstd{0.730}{0.002} & \meanstd{0.725}{0.000} & \meanstd{0.730}{0.002} & \meanstd{0.719}{0.002} & \meanstd{0.726}{0.001} \\
& GLPN-LLM & \meanstd{0.869}{0.004} & \meanstd{0.879}{0.006} & \meanstd{0.873}{0.006} & \meanstd{0.878}{0.005} & \meanstd{0.872}{0.001} & \meanstd{0.875}{0.005} \\
\midrule
\multirow{3}{*}{Ours}
& RAEGNet-E & \meanstd{0.889}{0.007} & \meanstd{0.901}{0.010} & \meanstd{0.879}{0.009} & \meanstd{0.906}{0.010} & \meanstd{0.969}{0.007} & \meanstd{0.898}{0.008} \\
& RAEGNet-B & \meanstd{0.905}{0.004} & \meanstd{0.912}{0.007} & \meanstd{0.893}{0.009} & \meanstd{0.919}{0.009} & \meanstd{0.980}{0.007} & \meanstd{0.911}{0.005} \\
& RAEGNet-A & \meanstd{0.913}{0.007} & \meanstd{0.923}{0.013} & \meanstd{0.899}{0.012} & \meanstd{0.924}{0.008} & \meanstd{0.985}{0.006} & \meanstd{0.919}{0.007} \\
\bottomrule
\end{tabular}
\end{table*}

\begin{table*}[t]
\centering
\small
\setlength{\tabcolsep}{5.5pt}
\renewcommand{\arraystretch}{1.08}
\caption{Accuracy across harm levels on SSS.}
\label{tab:harm_sss}
\begin{tabular}{clcccccc}
\toprule
Category & Method & 0.1 & 0.3 & 0.5 & 0.7 & 0.9 & Overall \\
\midrule
\multirow{4}{*}{w/o EK}
& CAFE & \meanstd{0.731}{0.009} & \meanstd{0.722}{0.010} & \meanstd{0.729}{0.011} & \meanstd{0.724}{0.010} & \meanstd{0.722}{0.007} & \meanstd{0.726}{0.009} \\
& MRML & \meanstd{0.752}{0.005} & \meanstd{0.762}{0.006} & \meanstd{0.756}{0.006} & \meanstd{0.761}{0.006} & \meanstd{0.756}{0.005} & \meanstd{0.758}{0.006} \\
& Event-Radar & \meanstd{0.755}{0.010} & \meanstd{0.748}{0.011} & \meanstd{0.760}{0.002} & \meanstd{0.751}{0.012} & \meanstd{0.752}{0.006} & \meanstd{0.754}{0.005} \\
& MSACA & \meanstd{0.757}{0.012} & \meanstd{0.768}{0.011} & \meanstd{0.760}{0.012} & \meanstd{0.765}{0.011} & \meanstd{0.759}{0.009} & \meanstd{0.762}{0.011} \\
\midrule
\multirow{3}{*}{w/ EK}
& KEHGNN-FD & \meanstd{0.680}{0.010} & \meanstd{0.668}{0.011} & \meanstd{0.677}{0.010} & \meanstd{0.671}{0.011} & \meanstd{0.674}{0.008} & \meanstd{0.674}{0.010} \\
& NSLM & \meanstd{0.775}{0.010} & \meanstd{0.785}{0.010} & \meanstd{0.778}{0.011} & \meanstd{0.784}{0.011} & \meanstd{0.780}{0.012} & \meanstd{0.781}{0.011} \\
& ERIC-FND & \meanstd{0.662}{0.009} & \meanstd{0.649}{0.009} & \meanstd{0.658}{0.010} & \meanstd{0.652}{0.009} & \meanstd{0.656}{0.007} & \meanstd{0.655}{0.009} \\
\midrule
\multirow{3}{*}{w/ LLM}
& Qwen2.5-VL & \meanstd{0.630}{0.001} & \meanstd{0.619}{0.001} & \meanstd{0.629}{0.001} & \meanstd{0.621}{0.001} & \meanstd{0.625}{0.002} & \meanstd{0.629}{0.005} \\
& InternVL2.5 & \meanstd{0.620}{0.005} & \meanstd{0.607}{0.004} & \meanstd{0.616}{0.004} & \meanstd{0.610}{0.006} & \meanstd{0.617}{0.005} & \meanstd{0.613}{0.005} \\
& GLPN-LLM & \meanstd{0.713}{0.008} & \meanstd{0.703}{0.006} & \meanstd{0.710}{0.008} & \meanstd{0.704}{0.007} & \meanstd{0.711}{0.006} & \meanstd{0.707}{0.007} \\
\midrule
\multirow{3}{*}{Ours}
& RAEGNet-E & \meanstd{0.811}{0.010} & \meanstd{0.808}{0.010} & \meanstd{0.779}{0.010} & \meanstd{0.818}{0.011} & \meanstd{0.853}{0.009} & \meanstd{0.808}{0.010} \\
& RAEGNet-B & \meanstd{0.842}{0.010} & \meanstd{0.840}{0.011} & \meanstd{0.809}{0.012} & \meanstd{0.844}{0.012} & \meanstd{0.878}{0.011} & \meanstd{0.837}{0.011} \\
& RAEGNet-A & \meanstd{0.847}{0.012} & \meanstd{0.845}{0.012} & \meanstd{0.815}{0.012} & \meanstd{0.851}{0.012} & \meanstd{0.888}{0.012} & \meanstd{0.843}{0.012} \\
\bottomrule
\end{tabular}
\end{table*}

\begin{table*}[t]
\centering
\small
\setlength{\tabcolsep}{6pt}
\renewcommand{\arraystretch}{1.08}
\caption{Direct evaluation of conditional-harm prediction.}
\label{tab:harm_prediction}
\begin{tabular}{llcc}
\toprule
Dataset & Predictor & MAE & RMSE \\
\midrule
\multirow{2}{*}{Weibo-21}
& Text-only MLP & \meanstd{0.103}{0.005} & \meanstd{0.137}{0.005} \\
& RAEGNet harm branch & \meanstd{0.076}{0.005} & \meanstd{0.103}{0.005} \\
\midrule
\multirow{2}{*}{Fakeddit}
& Text-only MLP & \meanstd{0.112}{0.005} & \meanstd{0.145}{0.004} \\
& RAEGNet harm branch & \meanstd{0.084}{0.006} & \meanstd{0.112}{0.006} \\
\midrule
\multirow{2}{*}{SSS}
& Text-only MLP & \meanstd{0.119}{0.008} & \meanstd{0.153}{0.008} \\
& RAEGNet harm branch & \meanstd{0.091}{0.007} & \meanstd{0.119}{0.006} \\
\bottomrule
\end{tabular}
\end{table*}

\begin{table*}[t]
\centering
\small
\setlength{\tabcolsep}{5.8pt}
\renewcommand{\arraystretch}{1.08}
\caption{Harm-matched evaluation. Held-out samples are balanced between real and fake news within each harm level in every fold.}
\label{tab:harm_matched_evaluation}
\begin{tabular}{ll l cccc}
\toprule
Dataset & Evaluation Set & Model & Accuracy & Macro F1 & Fake F1 & HHF F1 \\
\midrule
\multirow{4}{*}{Weibo-21}
& \multirow{2}{*}{Original}
& w/o Harm-aware Risk & \meanstd{0.915}{0.006} & \meanstd{0.915}{0.006} & \meanstd{0.914}{0.005} & \meanstd{0.946}{0.007} \\
& & Full Model & \meanstd{0.931}{0.009} & \meanstd{0.931}{0.009} & \meanstd{0.934}{0.009} & \meanstd{0.972}{0.006} \\
\cmidrule(lr){2-7}
& \multirow{2}{*}{Harm-matched}
& w/o Harm-aware Risk & \meanstd{0.902}{0.010} & \meanstd{0.902}{0.009} & \meanstd{0.903}{0.010} & \meanstd{0.928}{0.010} \\
& & Full Model & \meanstd{0.914}{0.010} & \meanstd{0.914}{0.010} & \meanstd{0.916}{0.010} & \meanstd{0.946}{0.010} \\
\midrule
\multirow{4}{*}{Fakeddit}
& \multirow{2}{*}{Original}
& w/o Harm-aware Risk & \meanstd{0.912}{0.010} & \meanstd{0.911}{0.010} & \meanstd{0.920}{0.010} & \meanstd{0.925}{0.011} \\
& & Full Model & \meanstd{0.919}{0.007} & \meanstd{0.918}{0.007} & \meanstd{0.926}{0.006} & \meanstd{0.947}{0.006} \\
\cmidrule(lr){2-7}
& \multirow{2}{*}{Harm-matched}
& w/o Harm-aware Risk & \meanstd{0.895}{0.010} & \meanstd{0.894}{0.010} & \meanstd{0.903}{0.010} & \meanstd{0.907}{0.010} \\
& & Full Model & \meanstd{0.904}{0.007} & \meanstd{0.903}{0.007} & \meanstd{0.912}{0.006} & \meanstd{0.927}{0.006} \\
\midrule
\multirow{4}{*}{SSS}
& \multirow{2}{*}{Original}
& w/o Harm-aware Risk & \meanstd{0.839}{0.010} & \meanstd{0.838}{0.010} & \meanstd{0.850}{0.010} & \meanstd{0.878}{0.010} \\
& & Full Model & \meanstd{0.843}{0.012} & \meanstd{0.841}{0.012} & \meanstd{0.860}{0.011} & \meanstd{0.896}{0.011} \\
\cmidrule(lr){2-7}
& \multirow{2}{*}{Harm-matched}
& w/o Harm-aware Risk & \meanstd{0.815}{0.006} & \meanstd{0.814}{0.007} & \meanstd{0.829}{0.006} & \meanstd{0.861}{0.006} \\
& & Full Model & \meanstd{0.825}{0.005} & \meanstd{0.823}{0.005} & \meanstd{0.839}{0.005} & \meanstd{0.879}{0.006} \\
\bottomrule
\end{tabular}
\end{table*}

\begin{table*}[t]
\centering
\small
\setlength{\tabcolsep}{5.2pt}
\renewcommand{\arraystretch}{1.08}
\caption{Controlled comparison of retrieval granularity and model architecture under the same source pool, filtering process, and evidence budget.}
\label{tab:controlled_retrieval_model}
\begin{tabular}{ll l cccc}
\toprule
Dataset & Retrieval Query & Backbone & Accuracy & Macro F1 & Fake F1 & HHF F1 \\
\midrule
\multirow{4}{*}{Weibo-21}
& \multirow{2}{*}{Entity-level}
& Evidence Fusion & \meanstd{0.889}{0.004} & \meanstd{0.889}{0.004} & \meanstd{0.891}{0.004} & \meanstd{0.932}{0.005} \\
& & RAEGNet & \meanstd{0.914}{0.005} & \meanstd{0.914}{0.005} & \meanstd{0.916}{0.005} & \meanstd{0.955}{0.006} \\
\cmidrule(lr){2-7}
& \multirow{2}{*}{Event-level}
& Evidence Fusion & \meanstd{0.907}{0.010} & \meanstd{0.907}{0.010} & \meanstd{0.910}{0.010} & \meanstd{0.946}{0.010} \\
& & RAEGNet & \meanstd{0.928}{0.005} & \meanstd{0.928}{0.005} & \meanstd{0.930}{0.004} & \meanstd{0.968}{0.003} \\
\midrule
\multirow{4}{*}{Fakeddit}
& \multirow{2}{*}{Entity-level}
& Evidence Fusion & \meanstd{0.872}{0.008} & \meanstd{0.870}{0.008} & \meanstd{0.884}{0.009} & \meanstd{0.912}{0.009} \\
& & RAEGNet & \meanstd{0.899}{0.012} & \meanstd{0.898}{0.011} & \meanstd{0.908}{0.012} & \meanstd{0.934}{0.011} \\
\cmidrule(lr){2-7}
& \multirow{2}{*}{Event-level}
& Evidence Fusion & \meanstd{0.892}{0.008} & \meanstd{0.890}{0.008} & \meanstd{0.903}{0.008} & \meanstd{0.925}{0.009} \\
& & RAEGNet & \meanstd{0.913}{0.005} & \meanstd{0.912}{0.005} & \meanstd{0.923}{0.005} & \meanstd{0.948}{0.005} \\
\midrule
\multirow{4}{*}{SSS}
& \multirow{2}{*}{Entity-level}
& Evidence Fusion & \meanstd{0.781}{0.004} & \meanstd{0.780}{0.004} & \meanstd{0.798}{0.004} & \meanstd{0.861}{0.004} \\
& & RAEGNet & \meanstd{0.809}{0.009} & \meanstd{0.808}{0.009} & \meanstd{0.823}{0.009} & \meanstd{0.879}{0.008} \\
\cmidrule(lr){2-7}
& \multirow{2}{*}{Event-level}
& Evidence Fusion & \meanstd{0.805}{0.009} & \meanstd{0.804}{0.009} & \meanstd{0.821}{0.010} & \meanstd{0.873}{0.009} \\
& & RAEGNet & \meanstd{0.836}{0.009} & \meanstd{0.835}{0.009} & \meanstd{0.852}{0.009} & \meanstd{0.896}{0.008} \\
\bottomrule
\end{tabular}
\end{table*}

\begin{table*}[t]
\centering
\small
\setlength{\tabcolsep}{5pt}
\renewcommand{\arraystretch}{1.08}
\caption{Effect of replacing the original external evidence of w/ EK baselines with ELERF.}
\label{tab:ek_event_evidence_transfer}
\begin{tabular}{lllccc}
\toprule
Dataset & Method & Evidence Source & Accuracy & Fake F1 & Real F1 \\
\midrule
\multirow{6}{*}{Weibo-21}
& \multirow{2}{*}{KEHGNN-FD} & Original & \meanstd{0.764}{0.005} & \meanstd{0.776}{0.004} & \meanstd{0.751}{0.005} \\
& & ELERF & \meanstd{0.801}{0.008} & \meanstd{0.812}{0.008} & \meanstd{0.789}{0.008} \\
\cmidrule(lr){2-6}
& \multirow{2}{*}{NSLM} & Original & \meanstd{0.878}{0.009} & \meanstd{0.885}{0.008} & \meanstd{0.870}{0.010} \\
& & ELERF & \meanstd{0.896}{0.005} & \meanstd{0.901}{0.005} & \meanstd{0.890}{0.004} \\
\cmidrule(lr){2-6}
& \multirow{2}{*}{ERIC-FND} & Original & \meanstd{0.844}{0.011} & \meanstd{0.857}{0.010} & \meanstd{0.828}{0.012} \\
& & ELERF & \meanstd{0.879}{0.011} & \meanstd{0.886}{0.011} & \meanstd{0.871}{0.010} \\
\midrule
\multirow{6}{*}{Fakeddit}
& \multirow{2}{*}{KEHGNN-FD} & Original & \meanstd{0.763}{0.010} & \meanstd{0.796}{0.008} & \meanstd{0.716}{0.012} \\
& & ELERF & \meanstd{0.792}{0.008} & \meanstd{0.823}{0.008} & \meanstd{0.748}{0.008} \\
\cmidrule(lr){2-6}
& \multirow{2}{*}{NSLM} & Original & \meanstd{0.850}{0.006} & \meanstd{0.867}{0.005} & \meanstd{0.828}{0.007} \\
& & ELERF & \meanstd{0.873}{0.008} & \meanstd{0.887}{0.008} & \meanstd{0.855}{0.008} \\
\cmidrule(lr){2-6}
& \multirow{2}{*}{ERIC-FND} & Original & \meanstd{0.794}{0.010} & \meanstd{0.828}{0.009} & \meanstd{0.744}{0.013} \\
& & ELERF & \meanstd{0.841}{0.004} & \meanstd{0.861}{0.004} & \meanstd{0.812}{0.004} \\
\midrule
\multirow{6}{*}{SSS}
& \multirow{2}{*}{KEHGNN-FD} & Original & \meanstd{0.674}{0.010} & \meanstd{0.698}{0.009} & \meanstd{0.644}{0.011} \\
& & ELERF & \meanstd{0.710}{0.009} & \meanstd{0.735}{0.009} & \meanstd{0.681}{0.010} \\
\cmidrule(lr){2-6}
& \multirow{2}{*}{NSLM} & Original & \meanstd{0.781}{0.010} & \meanstd{0.804}{0.009} & \meanstd{0.750}{0.012} \\
& & ELERF & \meanstd{0.806}{0.011} & \meanstd{0.826}{0.011} & \meanstd{0.782}{0.011} \\
\cmidrule(lr){2-6}
& \multirow{2}{*}{ERIC-FND} & Original & \meanstd{0.655}{0.009} & \meanstd{0.724}{0.007} & \meanstd{0.539}{0.013} \\
& & ELERF & \meanstd{0.717}{0.009} & \meanstd{0.758}{0.009} & \meanstd{0.660}{0.010} \\
\bottomrule
\end{tabular}
\end{table*}

\end{document}